\documentclass[11pt]{article}

\usepackage[preprint]{acl}

\usepackage{times}
\usepackage{latexsym}
\usepackage{url}
\usepackage{float}
\usepackage{booktabs}
\usepackage{graphicx}
\usepackage{multirow}
\usepackage{enumitem}
\usepackage{makecell}
\usepackage{amssymb}
\usepackage{wrapfig}
\usepackage{titletoc} 
\usepackage{hyperref} 
\usepackage{xcolor}
\usepackage{lineno}
\usepackage[most]{tcolorbox}
\usepackage{xargs}
\usepackage[colorinlistoftodos,prependcaption,textsize=tiny]{todonotes}
\definecolor{darkblue}{rgb}{0, 0, 0.5}
\usepackage{xcolor} 
\usepackage{xargs}  
\usepackage[colorinlistoftodos,textsize=tiny]{todonotes}
\usepackage{amsmath}
\newcommandx{\guofang}[2][1=]{\todo[linecolor=blue,backgroundcolor=blue!25,bordercolor=blue,#1]{#2}}

\usepackage[T1]{fontenc}

\usepackage[utf8]{inputenc}

\usepackage{microtype}

\usepackage{inconsolata}

\usepackage{graphicx}
\usepackage{fontawesome5}
\title{\faUserGraduate \ DeepInstructor: An Agentic AI Instructor for \\Experience-Driven Idea Evaluation}

\author{
  \textbf{Rongcan Pei\textsuperscript{1,*}},
  \textbf{Fang Guo\textsuperscript{2,*}},
  \textbf{Qinglin Qi\textsuperscript{2}},
  \textbf{Qi Zhu\textsuperscript{3}},
\\
  \textbf{Yun Luo\textsuperscript{4}},
  \textbf{Jianhao Yan\textsuperscript{4}},
  \textbf{Minjun Zhu\textsuperscript{2}},
  \textbf{Qiujie Xie \textsuperscript{2}},
\\
  \textbf{Dehong Zheng\textsuperscript{5}}, 
  \textbf{Yue Zhang\textsuperscript{2,\dag}},
\\
\\
  \textsuperscript{1}Tongji University,
  \textsuperscript{2}Westlake University,
  \textsuperscript{3}Zhejiang University,
  \textsuperscript{4}Shanghai AI Lab,
  \textsuperscript{5}Fudan University
\\
  \small{
    \textbf{Correspondence:} \href{prc@tongji.edu.cn}{prc@tongji.edu.cn}
  }
}

\begin{document}
\maketitle
\begingroup
\renewcommand\thefootnote{}
\footnotetext{* Equal contribution.}
\footnote{\dag Corresponding Author.}
\endgroup
\begin{abstract}

As automated scientific discovery advances, Large Language Models (LLMs) can now generate research ideas at an unprecedented scale, shifting the bottleneck from idea generation to idea evaluation. Existing evaluators mainly rely on parametric LLM knowledge or unstructured retrieval, producing judgments that lack the experience-grounded reasoning used by human instructors. To address this, we propose \textbf{DeepInstructor}, an agentic framework that formulates idea evaluation as reasoning over structured scholarly experience. DeepInstructor constructs an Experience Graph from 58,607 peer reviews and employs a ReAct-based agent to retrieve dimension-specific evidence for traceable evaluation. We further introduce \textbf{DeepInstruct}, a dataset with controlled pairwise comparisons across novelty, significance, and feasibility. Experiments show that DeepInstructor substantially outperforms existing baselines, improving Hit@1 and Hit@2 alignment with human judgments by 24.4\% and 29.7\%, respectively. Our findings suggest that scientific idea evaluation can be grounded in explicit reasoning over structured scholarly experience.
\footnote{Code, prompts, datasets and interview study results can be found on \href{https://anonymous.4open.science/r/ExpInstructor-5543/README.md}{Anonymous GitHub}.}
\end{abstract}

\section{Introduction}

The rapid advancement of AI enables both human researchers and emerging AI Scientist systems to generate research ideas at an unprecedented scale \citep{deepsci, zheng2025automation, zhang2025exploringLLMinsci, lu2026automationai, lu2024aiscientistfullyautomated}. 
As idea generation becomes increasingly cheap and automated \citep{scimon, uiucidea,huaweiidea}, evaluating idea quality remains costly and difficult, creating a growing imbalance in the scientific discovery process. 
Consequently, the bottleneck of scientific discovery is shifting from generating ideas to judging them \citep{Si2025IdeaExe, ScholarEval}, making robust idea evaluation a fundamental requirement for both AI Scientist systems and scalable human-AI collaboration \citep{zhang2025exploringLLMinsci,qiu2025aibench}.


Recent work on idea evaluation can be characterized along two orthogonal dimensions: what criteria to evaluate and how to evaluate.
Existing approaches vary widely in criteria—from single acceptability scores\citep{grapheval} to pairwise ranking and multi-dimensional rubrics—but it remains unclear whether these reflect how human instructors assess ideas in practice \citep{grapheval, Wen2025PredictingEA, ScholarEval, baek2025}. 
Methodologically, prompt-based approaches rely on parametric LLM knowledge and often misalign with human judgment\citep{Si2025IdeaExe}, while retrieval-based methods introduce noise due to irrelevant evidence from the content of papers \citep{ScholarEval}. Despite these differences, a shared limitation persists: \textbf{current evaluators fail to align with human expert judgment in either their criteria or their methodology.}


\begin{figure*}[t]
    \includegraphics[width=\textwidth]{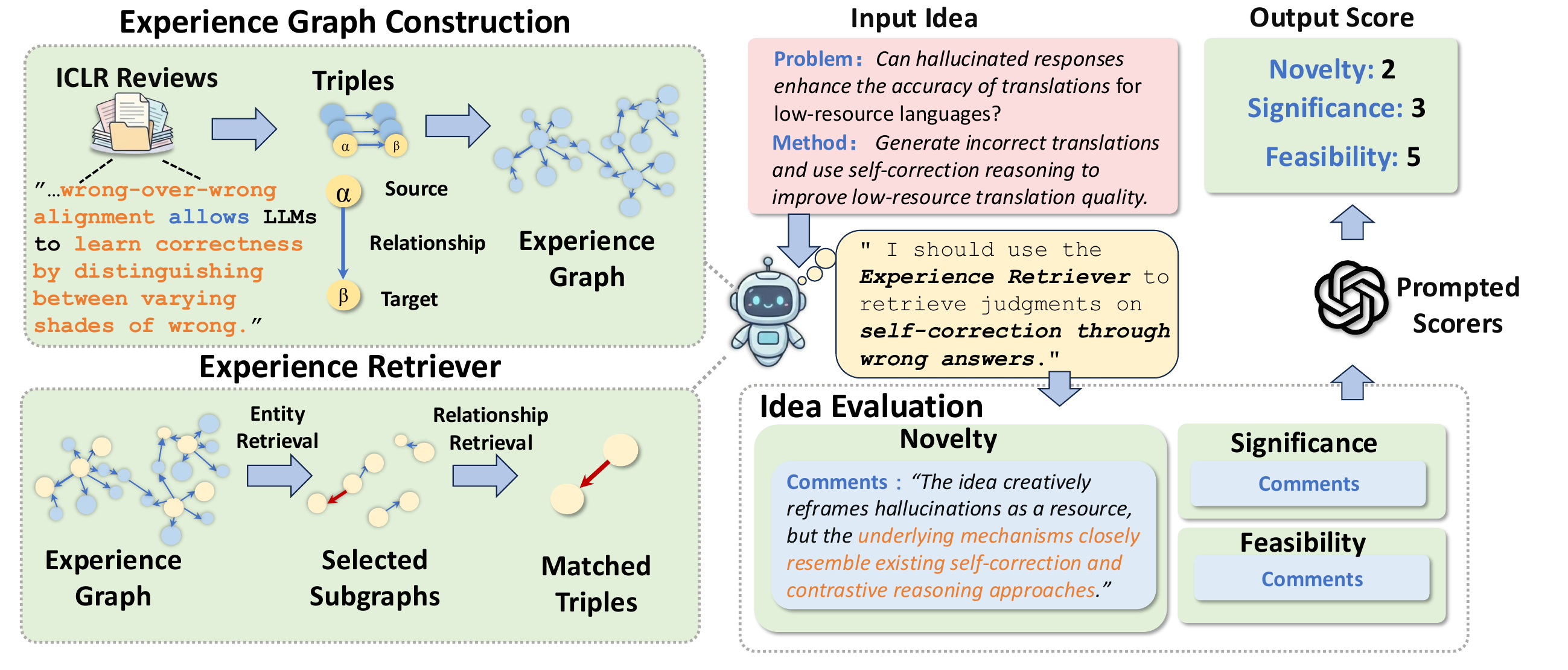}
    \caption{Framework of our proposed DeepInstructor. }
    \label{pipeline}
\end{figure*}

To better characterize human-aligned evaluation, we first conduct a pilot study with 15 Computer Science faculty members (see Appendix~\ref{appendix:pilot} for details). The study reveals two key insights: (1) \textbf{Novelty}, \textbf{Significance}, and \textbf{Feasibility} are the three criteria that instructors consistently prioritize, and (2) \textbf{the prior experience instructors rely on often takes the form of stance-bearing judgments toward existing techniques, algorithms, and tasks.} These findings suggest that idea evaluation is fundamentally an experience-grounded reasoning process, where judgments are formed through accumulated stances toward entities and their relations. We therefore conceptualize academic experience as a collection of such stance-bearing statements over conceptual entities and relations, providing a structured and human-aligned basis for evaluation.

For example, as shown in Figure~\ref{pipeline}, consider an idea proposing to improve low-resource machine translation by intentionally generating hallucinated incorrect translations and using them for self-correction. Rather than viewing the idea in isolation, experienced reviewers may relate it to existing paradigms, such as learning from incorrect answers and self-correction prompting. Such experience-grounded reasoning may lead reviewers to conclude that the idea recombines existing self-correction mechanisms rather than representing a fundamentally new paradigm.

Based on these insights, we propose \textbf{DeepInstructor}, an experience-driven framework for research idea evaluation that explicitly models scholarly experience and enables grounded, human-aligned reasoning. Our approach consists of three key components. First, we construct an \textbf{Experience Graph} from large-scale peer-review corpora with 58,607 cases, where entities and their evaluative relations encode accumulated scholarly experience as structured, stance-bearing knowledge. 
Second, we design an \textbf{Experience Retriever}. This graph-based retrieval module enables structured access to relevant experience by first identifying key entities in the input idea and then retrieving their associated evaluative relations conditioned on the target dimension. Third, we develop an \textbf{idea evaluation framework} based on the ReAct paradigm~\citep{react}, which integrates reasoning and retrieval to gather evidence and produce dimension-specific judgments dynamically. 

To enable controlled and systematic evaluation of research idea assessment, we further construct \textbf{DeepInstruct}, a benchmark derived from large-scale peer-review corpora containing nearly 20,000 ICLR submissions and reviews. By organizing papers along their task--method structure, we create controlled pairwise comparisons: ideas with similar tasks but different methods enable evaluation of novelty and feasibility, while those with similar methods across tasks support assessment of significance.

DeepInstructor is, to our knowledge, the first framework to model scholarly experience as structured evaluative relations and to leverage them for idea evaluation. We make three main contributions: (1) We propose DeepInstructor, an experience-driven evaluation paradigm that represents scholarly experience as a structured graph and enables dimension-specific retrieval for grounded idea assessment. (2) We introduce DeepInstruct, a principled benchmark construction pipeline that transforms peer-review corpora into controlled pairwise evaluation tasks. (3) We demonstrate that grounding evaluation in structured scholarly experience substantially improves alignment with human judgments, highlighting the potential of experience-grounded reasoning for scientific idea evaluation.

\section{Related Work}

\textbf{Automatic Research Idea Evaluation.}
Prior systems differ widely in evaluation criteria: DeepReview \citep{deepreview} offers reviewer-style judgments but lacks instructor-oriented dimensions; GraphEval \citep{grapheval} predicts only a single overall score for the entire paper without dimension-specific evaluation; ResearchAgent \citep{baek2025} uses 15 redundant dimensions; SciJudge\citep{qiu_taste} regards citation count as a metric; and \cite{canllm} evaluates along four axes. Methodologically, literature-based Retrieval-Augmented-Generation (RAG) tools \citep{ScholarEval} retrieve evidence but miss reviewers’ critical expertise, while review-based fine-tuning \citep{deepreview} and manual annotation \citep{canllm} remain costly. We instead extract reviewer experience directly and build an experience-grounded RAG agent for more actionable evaluations.

\textbf{Knowledge Graph Construction and RAG.}
LLM-based OpenIE increasingly produces schema-free triples \citep{openIE-survey}; Knowledge Graph Construction pipelines treat these as high-recall inputs that require normalization and linking \citep{IE-KG,LOKE}, which we apply when constructing our Experience Graph. Integrating KGs with RAG has further improved complex reasoning \citep{graphRAG,GRAG}, motivating our tailored retriever to efficiently locate reviewer experience.


\section{Methodology}

\paragraph{Problem Formulation.}
Given a research idea $I$ that consists of a problem formulation, a methodological sketch, and an experimental design, the idea evaluator $f$ is expected to generate structured evaluations along certain dimensions. Based on pilot interviews with domain experts, we focus on three key dimensions: \textit{Novelty}, which measures the originality of the idea relative to existing approaches; \textit{Significance}, which captures the importance and potential impact of the underlying problem; and \textit{Feasibility}, which reflects both the practicality of implementation and the expected effectiveness of the proposed method.

For each dimension, the evaluator outputs a scalar score reflecting its judgment. Formally, this process is defined as:
$f(I) \rightarrow \{s_d\}_{d \in \mathcal{D}},$
where $\mathcal{D} = \{\text{Novelty}, \text{Significance}, \text{Feasibility}\}$, and $s_d$ is the corresponding scalar score for each dimension.

\paragraph{Method Overview.}

\textbf{DeepInstructor} is an agentic framework for research idea evaluation that leverages structured experience distilled from peer reviews. To encode and reuse historical judgements and experience, DeepInstructor consists of three components: (1) an \textbf{Experience Graph} that stores entities and their evaluative relations as externalized scholarly experience; (2) an \textbf{Experience Retriever} that performs entity- and relation-aware retrieval over the graph; and (3) an \textbf{idea evaluation framework} based on ReAct that integrates reasoning and retrieval to produce dimension-specific, experience-grounded judgments. The overall pipeline is illustrated in Figure~\ref{pipeline}.

\subsection{Experience Graph Construction}
Peer reviews encode rich evaluative knowledge grounded in prior research experience. To capture and reuse such historical knowledge, we construct an \textbf{Experience Graph} from ICLR 2024 and ICLR 2025 reviews.

\paragraph{Extracting Triples.}
We treat each review as a fundamental unit for experience extraction. Using a prompted LLM, we extract scientific entities (nodes) and identify their precise evaluative or methodological relations (edges), accompanied by the original textual evidence.
For example, as shown in Figure~\ref{pipeline}, given the review statement, ``wrong-over-wrong alignment allows LLMs to learn correctness by distinguishing between varying shades of wrong,'' the model extracts two entities (\textit{wrong-over-wrong alignment ($\alpha$)}, \textit{learning from incorrect answers ($\beta$)}), connected by the relation \textit{enables}. In this way, stance-bearing scholarly judgments are transformed into structured experience triples.
Finally, we extract 416,330 triplets in total.

\begin{figure}
    \includegraphics[width=\linewidth]{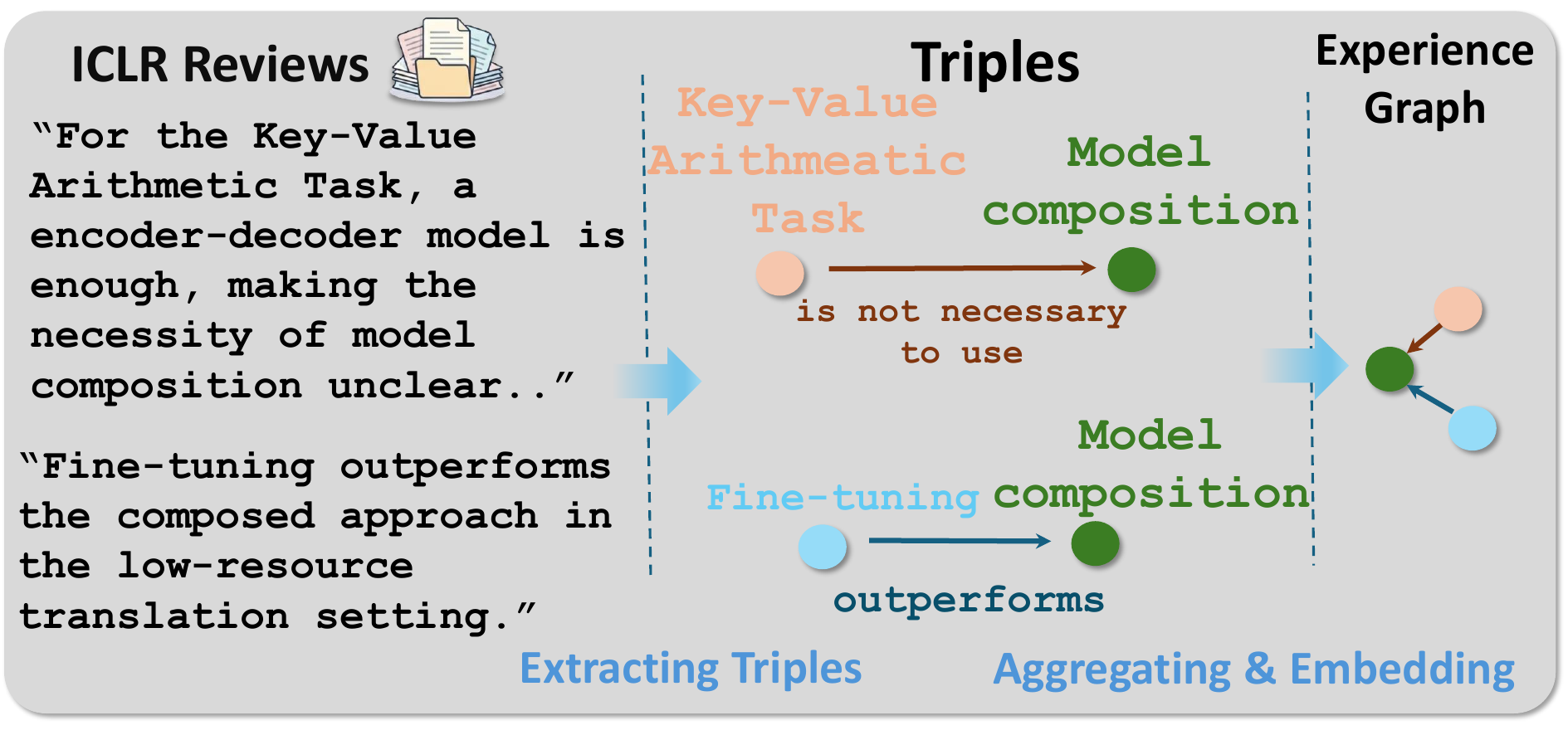}
    \caption{Examples of Triplets Extraction.}
    \label{fig:tri}
\end{figure}

\paragraph{Aggregating and Embedding.}
We aggregate extracted triples into a unified Experience Graph by merging identical nodes, transforming scattered reviews into subgraphs that capture both positive and negative historical evidence. Finally, we compute embeddings for all nodes and edges to enable semantic retrieval, establishing this graph as DeepInstructor's external memory. Formally, we define the Experience Graph as a directed labeled graph:
$\mathcal{G} = (\mathcal{V}, \mathcal{E}),$ where $\mathcal{V}$ denotes the set of entities and $\mathcal{E}$ denotes the set of relational edges. Each edge is represented as:
$e = (v_i, r, v_j, t) \in \mathcal{E},$
where $v_i, v_j \in \mathcal{V}$ are scientific entities after aggregating, $r$ is the extracted relation , and $t$ is the associated textual evidence.

Figure~\ref{fig:tri} shows more specific examples of how a graph is built from reviews. For detailed statistics of the constructed Experience Graph and the corresponding graph coverage evaluation, please refer to Appendix~\ref{triplets}.

\subsection{Experience Retriever}

We design a graph-based retrieval tool called Experience Retriever. As illustrated in Figure~\ref{pipeline}, the retriever is designed to mimic how human experts recall prior experience: they first identify the concepts most relevant to the current idea, and then focus on the particular evaluative relations that matter for the target judgment. Following this intuition, our retriever operates in two stages: \emph{knowledge entity retrieval} and \emph{relationship retrieval}.

Formally, we define a retrieval operator:
$\mathcal{R}: (K, K_c, R, R_c) \mapsto \mathcal{E}',$
where \(K\) denotes the knowledge entity query, \(K_c\) is the number of retrieved entities, \(R\) is the relationship query, and \(R_c\) is the number of returned experience edges. The output \(\mathcal{E}' \subseteq \mathcal{E}\) is a set of relevant edges in the Experience Graph. Given this input, the retriever operates in the following two steps: 

\paragraph{Step 1: Knowledge Entity Retrieval.}
The first step aims to identify graph nodes that are semantically related to the query concept \(K\). Embedding-based retrieval is adopted here to guarantee semantic alignment across different surface forms, enabling robust matching of conceptually similar entities.

Let \(\phi_v(\cdot)\) denote the embedding function for entity nodes. Given the entity query \(K\), we retrieve the top-\(K_c\) most similar entities:
$\mathcal{V}_K
=
\operatorname{Top}\text{-}K_c
\Big(
\{
\mathrm{sim}(\phi_v(K), \phi_v(v)) \mid v \in \mathcal{V}
\}
\Big),$
where \(\mathrm{sim}(\cdot,\cdot)\) is cosine similarity. These retrieved entities define the candidate retrieval space. Specifically, we collect all edges incident to any retrieved entity:
$
\mathcal{E}_{\mathrm{cand}}
=
\{\, e \in \mathcal{E} \mid \exists v \in \mathcal{V}_K,\; v \in e \,\}.
$

\paragraph{Step 2: Relationship Retrieval.}
After identifying the relevant entities, the second step focuses on selecting the relations that best match the current evaluative need. By matching the edges against a relationship query, this step effectively filters out irrelevant connections, ensuring that only the historical evidence tailored to the current evaluation dimension is retrieved.

Let \(\phi_r(\cdot)\) denote the embedding function for relation texts. For each candidate edge
$e = (v_i, r_e, v_j, x) \in \mathcal{E}_{\mathrm{cand}}, $
we compute its relevance to the relationship query \(R\) by matching \(R\) with the edge relation \(r_e\). We then retrieve the top-\(R_c\) most relevant candidate edges:
$\mathcal{E}_{R}
=
\operatorname{Top}\text{-}R_c
\Big(
\{
\mathrm{sim}(\phi_r(R), \phi_r(r_e)) \mid e \in \mathcal{E}_{\mathrm{cand}}
\}
\Big).$ The final retrieved evidence is given by
$ \mathcal{R}(K, K_c, R, R_c) = \mathcal{E}_{R},$
where each edge in $\mathcal{E}_{R}$ is associated with its supporting textual evidence $t$. This two-stage design decouples concept matching from relation matching, enabling dimension-specific retrieval for different evaluation criteria.


For example, as shown in Figure~\ref{pipeline}, given the query ``self-correction through wrong answers,'' the retriever first identifies the most relevant entities (e.g., $\langle \alpha, \Omega, \dots \rangle$) from the Experience Graph. It then explores the associated subgraphs to retrieve relations and scholarly judgments that best match the retrieval intent, such as prior evaluations of learning from incorrect answers or self-correction mechanisms. The retriever ultimately returns the most relevant experience triples to support downstream idea evaluation.

\subsection{Idea Evaluation Process}
Equipped with the Experience Graph as external memory and the Experience Retriever as a structured retrieval operator, we develop an agentic evaluation framework that dynamically grounds idea assessment in retrieved scholarly experience.  As illustrated in Figure~\ref{pipeline}, when evaluating the novelty of the translation hallucination idea, the agent retrieves experiences related to learning from incorrect answers and self-correction prompting, and synthesizes them into a grounded evaluative comment.

Given a research idea \(I\) and the dimension-wise prompt, DeepInstructor performs agentic retrieval and reasoning over the Experience Graph to construct dimension-specific comments. Conditioned on the comments, the scorer model produces a scalar score \(s_d\) for each evaluation dimension \(d \in \mathcal{D}\), reflecting its judgment grounded in accumulated reviewer experience.

\subsubsection{Agentic Retrieval and Reasoning}



To effectively leverage the Experience Retriever, we instantiate DeepInstructor as a ReAct-style agent \citep{react} that interleaves reasoning with tool use. Instead of following a fixed retrieval pipeline, the agent autonomously decides when to retrieve, which knowledge entities to query, and which evaluative relations are relevant to the current judgment.

For each evaluation dimension, the agent iteratively queries the Experience Graph for key concepts, accumulating historical evidence in its working memory until sufficient for evaluation. As illustrated in Figure~\ref{pipeline}, given the idea ``\textit{Generate incorrect translations and use self-correction reasoning...}'', DeepInstructor first queries the entity ``\textit{self-correction through wrong answers}'', and then queries related concepts such as ``\textit{machine translation}''.

The final retrieved evidence is then synthesized into a dimension-specific comment: $
c_d = LLM_{CG}(I, d, M_d),$
where \(M_d = \{ t \mid e \in \mathcal{E}_R \}\) denotes the set of textual evidence $t$ associated with the retrieved edges for dimension \(d\), and $LLM_{CG}$ denotes the LLM-based comment generation function.

\subsubsection{Score Prediction via LLM Scorer}
\label{llmscorer}

Directly prompting an LLM to output scores often leads to poorly calibrated predictions and highly compressed score distributions. We find an effective method to solve the issue--decouple qualitative judgment from quantitative scoring. Instead of predicting the score directly from the idea, we first generate the textual comment \(c_d\), and then map it to a scalar score using a separate dimension-specific, prompted with few-shot examples scorer:
$s_d = LLM_{score}(c_d),$
where $LLM_{score}$ is an LLM scorer instantiated with few-shot exemplars for dimension \(d\). It is proven to have good accuracy on experiment datasets in Appendix~\ref{sec:scorer}.


\section{Experimental Setup}

\subsection{Datasets}

We evaluate DeepInstructor on two datasets. The first golden evaluation dataset is drawn from a high-quality human annotations dataset~\citep{canllm}, which contains original research ideas (authored by humans or generated by AI for the NLP domain), human comments on each idea's novelty, significance, and feasibility (including implementability and expected effectiveness), together with corresponding numeric scores (0--10). We refer to this dataset as \textbf{ExpertEval}. ExpertEval includes 147 ideas with the expert evaluations and scores of the 3 dimensions.

To enable large-scale and fine-grained evaluation, we further construct \textbf{DeepInstruct}, a dataset derived from ICLR 2026 peer reviews, as they provide reliable evaluation signals. First, we cluster papers sharing identical tasks or methods. Based on peer-review evidence, we then grade each paper into high, medium, or low tiers across three evaluation dimensions(Novelty, Significance, and Feasibility). Next, we pair clustered papers with contrasting grades. 
Finally, we reformat all papers into a standardized research idea representation. \textbf{This design supports both single-paper level-based evaluation and pairwise comparison between contrasting papers.}
This pipeline is driven by LLMs and validated by human experts. The resulting dataset contains 398 ideas along with evaluation and 289 pairwise comparisons. Full construction details are provided in Appendix~\ref{appendix:dataset}.

 

\subsection{Evaluation Metrics}
We evaluate model performance under two settings, corresponding to the two datasets used in our experiments.

For ExpertEval, we focus on measuring alignment between model predictions and human judgments. 
Due to natural inter-rater variance, minor absolute deviations in review scores typically indicate practical equivalence rather than disagreement. However, standard correlation coefficients assume strict linearity and suffer from range restriction, making them ill-suited for the ordinal and clustered nature of human grading. By analyzing the score distribution features of ICLR 2026 reviews, we conclude that \textbf{Hit@1} and \textbf{Hit@2} are superior metrics compared to the Spearman correlation. 
We detail this proof in Appendix~\ref{sec:proof}.

For feasibility evaluation, scalar scores alone cannot capture whether a model identifies the same implementation risks as human experts. We therefore additionally report \textbf{Concern-level coverage} and \textbf{precision}, which measure whether generated feasibility concerns match human-annotated concerns. The detailed matching protocol is provided in Appendix~\ref{deepintrcut:metrics}.

For DeepInstruct, we evaluate whether models can distinguish ideas of different quality under controlled comparisons. We report \textbf{Pairwise Accuracy}, which measures whether the model correctly ranks two ideas with different quality levels, and \textbf{Level Accuracy}, which measures whether the model assigns each idea to the correct quality tier. Following our dataset labels, scores of 1--4 are classified as low, 5--6 as medium, and 7--9 as high.

\subsection{Compared Methods}
We compare our method against a diverse set of baselines, including a prior paper evaluation model, an RAG approach, and strong commercial LLMs:
\begin{itemize}
    \item \textbf{DeepReview}: A specialized model fine-tuned on a large-scale corpus of ICLR peer reviews \citep{deepreview}. Although it is designed for full-paper assessment, it can be adapted for idea evaluation by providing only a problem formulation and a methodological sketch as input.
    \item \textbf{RAG}: A baseline that replaces our structured graph-based retrieval with passage-level embedding retrieval. A similar implementation is used in ScholarEval \citep{ScholarEval}.
    \item \textbf{Commercial LLMs}: Prior work has shown that commercial LLMs can serve as strong baselines for scientific idea evaluation \citep{ScholarEval, qiu_taste, Si2025IdeaExe}. We include three representative commercial LLMs: \textit{ChatGPT-4o-mini}, \textit{ChatGPT-4o}, and \textit{DeepSeek-V4-Pro}, which differ in reasoning capability and parametric knowledge coverage.  In addition, these models are also used as the backbone models of DeepInstructor, enabling controlled comparisons between standalone LLM evaluation and our experience-grounded framework under the same underlying model.
    
\end{itemize}

\subsection{Implementation Details}

For DeepInstructor, we use ChatGPT-4o-mini for graph construction and adopt multiple backbone models, including ChatGPT-4o-mini, ChatGPT-4o, and DeepSeek-V4-Pro, for agentic reasoning ($LLM_{CG}$), and score generation ($LLM_{score}$). 
The Experience Graph is constructed from peer reviews of ICLR 2024 and 2025. For the Experience Retriever, we encode both entities and relations using BGE-M3 \citep{bgem3}, with retrieval sizes $K_c = 5$ and $R_c = 5$.  For idea evaluation, scores are normalized to a 10-point scale.  For RAG, we use ChatGPT-4o as the backbone model and retrieve the top-10 most relevant review sentences using the same embedding model (BGE-M3) for fair comparison and directly condition the LLM on the retrieved text, without structured retrieval or agentic reasoning. The temperature of all LLMs is set to 0.1. Further details are provided in Appendix~\ref{sec: Experimental Details}. 

\section{Results}

For experimental results, we first present the main evaluation results on two datasets, followed by an ablation study on different graph sizes and an AI-in-the-loop experiment to see whether the proposed framework can enhance the idea generator with comments from our proposed model.

\subsection{Main Results}

\subsubsection{Alignment Experiment on ExpertEval}
\begin{table*}[ht]
\centering
\caption{Performance comparison across Novelty, Significance, and Feasibility evaluation. 
We report Hit@$\pm 1$ and Hit@$\pm 2$ for scalar scores, and mean coverage and precision for feasibility concerns. $Ours$ indicates the proposed DeepInstructor Model. Best and second-best results are highlighted in \textbf{bold} and \underline{underlined}, respectively.}
\label{tab:combined_performance}
\resizebox{\textwidth}{!}{
\begin{tabular}{lcccccccc}
\toprule
\multirow{2}{*}{Model} & \multicolumn{2}{c}{Novelty} & \multicolumn{2}{c}{Significance} & \multicolumn{2}{c}{Feasibility Score} & \multicolumn{2}{c}{Feasibility Concerns} \\
\cmidrule(lr){2-3} \cmidrule(lr){4-5} \cmidrule(lr){6-7} \cmidrule(lr){8-9}
 & Hit@$\pm 1$ & Hit@$\pm 2$ & Hit@$\pm 1$ & Hit@$\pm 2$ & Hit@$\pm 1$ & Hit@$\pm 2$ & Mean Cov. & Mean Prec. \\
\midrule
$RAG$ & 0.452 & 0.681 & 0.000 & 0.133 & 0.681 & 0.933 & \underline{0.579} & 0.310 \\
DeepReview & 0.504 & 0.741 & 0.044 & 0.207 & 0.667 & 0.895 & 0.007 & 0.000 \\
ChatGPT-4o & 0.089 & 0.215 & 0.000 & 0.133 & 0.652 & 0.926 & 0.575 & 0.305 \\
ChatGPT-4o-mini & 0.044 & 0.141 & 0.000 & 0.141 & 0.719 & 0.970 & 0.440& 0.246 \\
DeepSeek-V4-Pro & 0.378 & 0.548 & 0.044 & 0.200 & 0.385 & 0.682 & 0.522 & \textbf{0.391} \\
\midrule
$Ours_{4o-mini}$ & 0.515 & 0.746 & 0.030 & 0.207 & \textbf{0.748} & \textbf{0.978} & 0.496 & 0.261 \\
$Ours_{4o}$ & \underline{0.622} & \textbf{0.867} & \underline{0.126} & \underline{0.356} & \underline{0.726} & \underline{0.941} & \textbf{0.640} & \underline{0.362} \\
$Ours_{DeepSeek}$ & \textbf{0.649} & \underline{0.863} & \textbf{0.207} & \textbf{0.422} & 0.585 & 0.896 & 0.534 & 0.355 \\
\bottomrule
\end{tabular}
}
\end{table*}

Table~\ref{tab:combined_performance} summarizes the evaluation results across all dimensions. We highlight several key observations: 

(1) \textbf{DeepInstructor consistently outperforms standalone LLM evaluators.} 
When using the same backbone model, DeepInstructor achieves substantial improvements over the corresponding standalone commercial models. 
For example, compared to ChatGPT-4o, DeepInstructor with ChatGPT-4o as backbone improves the average Hit@1 and Hit@2 by \textbf{24.4\%} and \textbf{29.7\%}, respectively. 
Notably, all standalone commercial LLMs achieve extremely low performance on Significance evaluation. Manual inspection suggests that these models tend to assign overly positive significance judgments to most ideas, making it difficult to distinguish genuinely impactful problems from marginal or incremental ones(see Appendix~\ref{case_study_sig} for a representative example). In contrast, grounding evaluation in peer-review experience enables DeepInstructor to better identify low-significance ideas, leading to substantially improved performance on this challenging dimension. 

(2) \textbf{Compared to RAG, DeepInstructor provides more effective evaluation capabilities.} 
While RAG achieves moderate performance on Novelty, it performs extremely poorly on Significance, with near-zero Hit rates. This suggests that conventional retrieval fails to capture the evaluative knowledge required for significance assessment. In contrast, DeepInstructor substantially improves performance on all three dimensions, highlighting the importance of structured experience modeling and relation-aware retrieval for evaluative reasoning (see Appendix~\ref{case-sd} for a representative example).

(3) \textbf{The framework generalizes across backbone models and reduces reliance on model scale.} 
Even with a weaker backbone (ChatGPT-4o-mini), DeepInstructor outperforms stronger baselines such as DeepReview and DeepSeek-V4-Pro on multiple dimensions. 
Interestingly, stronger general-purpose reasoning capability does not always translate to better idea evaluation performance. For example, DeepSeek-V4-Pro underperforms ChatGPT-4o and ChatGPT-4o-mini on feasibility evaluation despite being a stronger reasoning model overall, suggesting that research idea evaluation exhibits model-specific behavioral differences beyond raw model capability. Nevertheless, pairing DeepSeek-V4-Pro with DeepInstructor still leads to substantial improvements, further demonstrating the robustness of the proposed framework across heterogeneous backbones.

To better explain these performance differences, we provide detailed case studies on novelty and significance evaluation, as well as a walkthrough example illustrating how the DeepInstructor agent performs step-by-step reasoning in Appendix~\ref{case-sd}. The case studies show that DeepInstructor can retrieve both positive and skeptical reviewer judgments from historical peer reviews, enabling more calibrated and experience-grounded evaluation compared to standalone LLMs and RAG systems.

\subsubsection{Alignment Experiment on DeepInstruct}

\begin{table}[t]
\centering
\caption{Level Accuracy and Pairwise Comparison Accuracy on DeepInstruct. Best results are in \textbf{bold}, and second-best are \underline{underlined}. 4o-mini = ChatGPT-4o-mini; DS = DeepSeek-V4-Pro}
\label{tab:accuracy_summary}
\setlength{\tabcolsep}{3.5pt} 
\resizebox{\columnwidth}{!}{
\begin{tabular}{lccc}
\toprule
\textbf{Model} & \makecell{\textbf{Novelty (\%)} \\ \textit{Pair. / Lev.}} & \makecell{\textbf{Feasibility (\%)} \\ \textit{Pair. / Lev.}} & \makecell{\textbf{Significance (\%)} \\ \textit{Pair. / Lev.}} \\
\midrule
RAG & \underline{46.15} / 36.32 & 24.41 / 48.43 & 13.33 / 20.00 \\
4o-mini & 41.03 / 23.50 & 22.83 / 45.67 & 23.33 / 25.00 \\
DS & 33.33 / 40.31 & \textbf{42.64} / 15.50 & 30.00 / \underline{48.33} \\
\midrule
$\text{Ours}_{4o\text{-mini}}$ & \textbf{47.86} / \textbf{43.16} & 25.20 / \textbf{52.36} & \underline{36.67} / 26.67 \\
\quad $\Delta$ & \textcolor{blue}{+7\%} / \textcolor{blue}{+20\%} & \textcolor{blue}{+2\%} / \textcolor{blue}{+7\%} & \textcolor{blue}{+13\%} / \textcolor{blue}{+2\%} \\
\addlinespace 
$\text{Ours}_{\text{DS}}$ & 37.34 / \underline{42.58} & \underline{39.46} / \underline{50.39} & \textbf{40.00} / \textbf{49.11} \\
\quad $\Delta$ & \textcolor{blue}{+4\%} / \textcolor{blue}{+2\%} & \textcolor{red}{-3\%} / \textcolor{blue}{+35\%} & \textcolor{blue}{+10\%} / \textcolor{blue}{+1\%} \\
\bottomrule
\end{tabular}
}
\end{table}
Table~\ref{tab:accuracy_summary} presents the accuracy of classifying ideas into the correct levels, as well as pairwise comparison accuracy based on these levels. The results provide two additional insights that complement the findings on ExpertEval:

(1) \textbf{Robust performance under alternative evaluation protocols.}
DeepInstructor consistently improves Level accuracy while remaining competitive on Pairwise accuracy across different evaluation dimensions, demonstrating stronger calibration of absolute judgment. Compared to conventional RAG, DeepInstructor achieves substantially better Level accuracy, suggesting that passage-level retrieval alone is insufficient for reliable idea evaluation. Instead, grounding evaluation in structured scholarly experience enables more calibrated and reliable judgments. 

(2) \textbf{Structured experience improves score calibration.}
Although stronger commercial LLMs with more recent knowledge (e.g., DeepSeek-V4-Pro) achieve competitive Pairwise accuracy on certain dimensions, they still struggle with calibrated absolute scoring, leading to poor Level accuracy. DeepInstructor mitigates this issue by grounding judgments in structured scholarly experience, resulting in substantially improved calibration (e.g., +35\% Level accuracy on feasibility), even with a minor trade-off in Pairwise performance.

Overall, these results suggest that the gains of DeepInstructor are not solely explained by backbone model scale or newer parametric knowledge, but are closely related to the proposed experience-grounded retrieval and reasoning framework. The consistent improvements observed across both evaluation datasets further support the robustness of the proposed paradigm.

\subsection{Analysis of Experience Coverage}

To investigate the impact of experience coverage on model performance, we construct Experience Graphs using data from different years (2024 vs.\ 2025) and compare them with the full-coverage setting (both 2024 and 2025).

\begin{figure}[htbp]
\includegraphics[width=\linewidth]{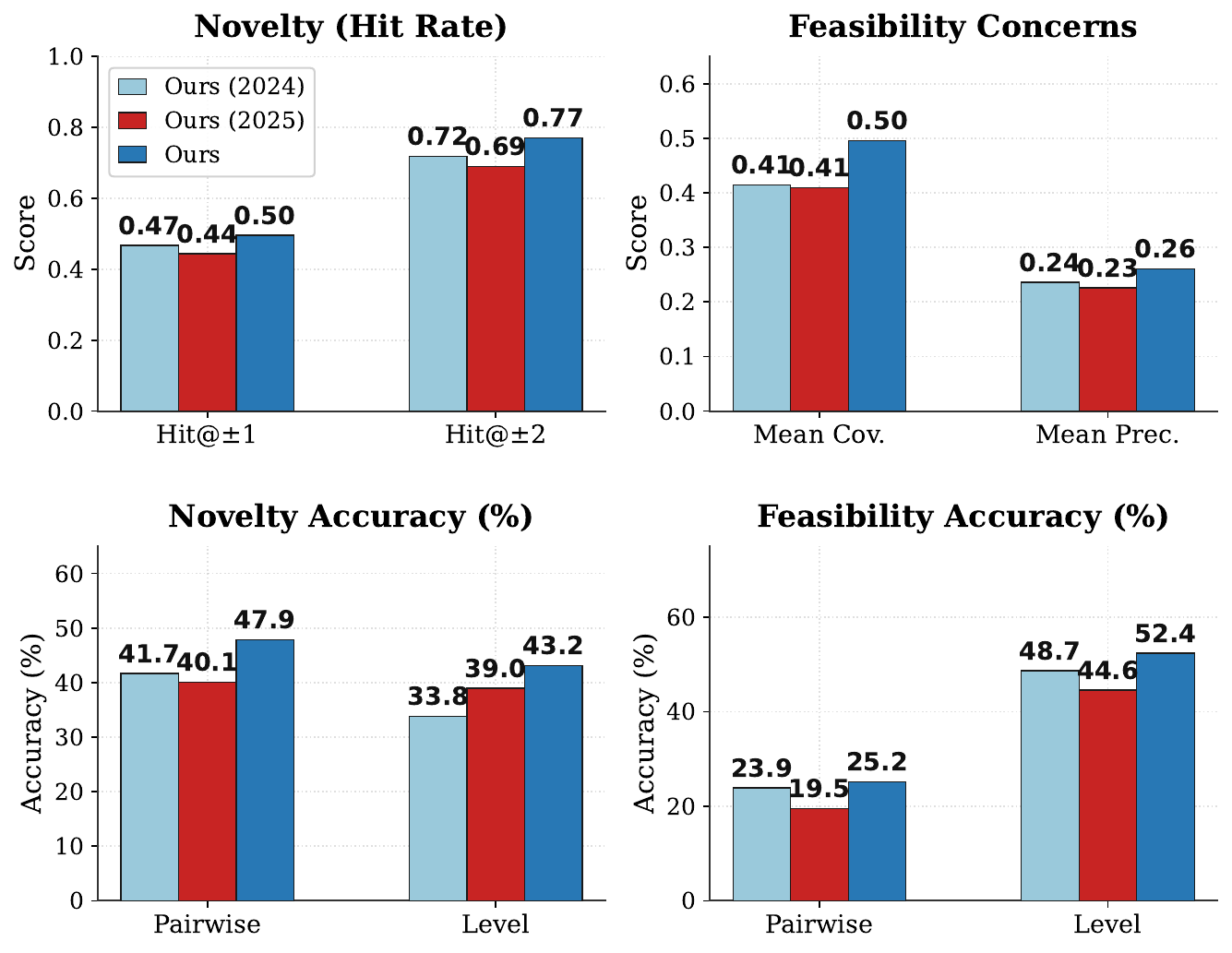}
\caption{Ablation study on the effect of experience coverage using different subsets of the Experience Graph. 
$Ours$ denotes the full-coverage setting, while $Ours_{2024}$ and $Ours_{2025}$ use year-specific data only.}
\label{tab:ours_main_result}

\end{figure}

As shown in Figure~\ref{tab:ours_main_result}, reducing the temporal coverage of experience leads to consistent performance degradation across both score-based and structure-aware metrics. In particular, using only 2025 data results in noticeable drops in Novelty alignment (Hit@$\pm1$: 0.444 vs.\ 0.496) and feasibility reasoning quality (coverage: 0.4092 vs.\ 0.4955), suggesting that limited experience fails to provide sufficient evaluative evidence. Interestingly, the 2024-only graph performs more competitively than the 2025-only graph on several metrics, indicating that performance is not solely determined by recency but also by the diversity and completeness of accumulated experience. This highlights that effective idea evaluation requires broad and well-distributed historical knowledge, rather than narrowly scoped or temporally localized evidence.
Overall, these results demonstrate that the effectiveness of DeepInstructor is closely tied to the coverage and richness of the Experience Graph, further supporting our design choice of leveraging large-scale, aggregated scholarly experience for robust evaluation.

\subsection{AI-in-the-loop Experiment}

To study the role of evaluative feedback in iterative idea refinement, we design an AI-in-the-loop experiment to simulate a closed-loop research iteration process. Starting from an initial idea $R_0$, the idea is iteratively refined over 3 rounds. At each iteration, the generator updates the current idea based on evaluation comments produced for the previous version. Specifically, we use the dimension-specific comments $c_d = LLM_{CG}(I, d, M_d)$ generated by DeepInstructor as structured evaluative feedback, and compare this setting against a generic prompt-based feedback baseline (``Please improve your idea''). To examine the effects of both feedback quality and generator capability, we conduct experiments using ChatGPT-4o and ChatGPT-4o-mini as generators. All experiments are conducted on the ExpertEval dataset.

\begin{table}[t]
\centering
\caption{Average overall feedback score improvements ($\bar{\Delta}$) across different methods.}
\label{tab:feedback_scores_simple}
\resizebox{\columnwidth}{!}{ 
\begin{tabular}{lc} 
\toprule
\textbf{Method} & \textbf{$\bar{\Delta}$} \\
\midrule
ChatGPT-4o-mini prompt feedback     & +0.068 \\
ChatGPT-4o-mini Instructor feedback & +0.121 \\
ChatGPT-4o Instructor feedback      & +0.203 \\
\bottomrule
\end{tabular}
}
\end{table}

Results in Table~\ref{tab:feedback_scores_simple} show that DeepInstructor consistently yields larger improvements than generic feedback, indicating that structured, experience-grounded critiques provide more effective guidance. Second, while stronger models (ChatGPT-4o) can better utilize high-quality feedback, weaker ones (ChatGPT-4o-mini) benefit less. This reveals a key interaction—high-quality feedback alone is insufficient without a capable generator to leverage it.

\section{Conclusion}

We present \textbf{DeepInstructor}, an experience-driven, agentic framework for evaluating research ideas that grounds judgments in structured scholarly experience. Experiments show that DeepInstructor achieves strong alignment with human evaluations and consistently outperforms existing baselines. More broadly, our work suggests that scholarly experience can be externalized and reused as a structured resource for AI reasoning. In future work, we plan to extend the experience source beyond peer reviews to broader scientific resources, such as published papers and scientific editorials.

\section{Limitations}
Although DeepInstructor substantially improves significance evaluation by grounding judgments in reviewer experience, the overall performance remains limited, suggesting that significance assessment is still an open challenge for experience-grounded idea evaluation. We leave improving significance-oriented retrieval and reasoning as important future work.

In addition, the current Experience Graph is constructed from only two years of open peer reviews in computer science, which limits both the coverage and diversity of evaluative experience. Extending DeepInstructor to other scientific domains may require incorporating additional experience sources beyond open reviews, since publicly available peer-review data in many fields remains limited.

\bibliography{custom}

\clearpage

\appendix
\section*{Appendix Catalog}
\startcontents[appendix]
\printcontents[appendix]{l}{1}{\setcounter{tocdepth}{2}}
\vspace{1em} 
\clearpage

\section{Pilot Interview Study}
\label{appendix:pilot}
To understand how human instructors evaluate research ideas and what forms of scholarly experience they rely on, we conducted a semi-structured pilot interview study with 15 Computer Science faculty members across machine learning, natural language processing, systems, and human–computer interaction. All participants had at least five years of advising experience and had supervised multiple student research projects.

Interviews were conducted remotely and lasted 20–30 minutes. We adopted a semi-structured protocol consisting of four themes: (1) criteria instructors consider when evaluating student ideas, (2) examples of strong and weak ideas, (3) the types of prior scholarly experience drawn upon during evaluation, and (4) the capabilities expected of an AI evaluator. The protocol encourages both structured reflection and open-ended elaboration, allowing us to capture both high-level evaluation criteria and rich, stance-bearing judgments derived from instructors’ accumulated research experience.

All interviews were audio-recorded, transcribed, and analyzed using open coding by two independent annotators. Through iterative thematic analysis, we identified recurring evaluation criteria and extracted representative experience-based statements, which subsequently informed the design of the Experience Graph and the dimension-specific evaluation strategies used in DeepInstructor.

\subsection{Pilot Interview Protocol}

We list below the full interview protocol used during the study.

\textbf{Theme A: Evaluation Criteria}

\begin{enumerate}[label=\textbf{Q\arabic*.}]
\item \textbf{Criteria for Evaluating Research Ideas} \\
When evaluating a new research idea from a student, which criteria do you typically consider?  
Please describe all aspects you examine when forming an initial judgment.  
(You may include considerations such as novelty, significance, feasibility, technical soundness, etc., but please feel free to answer in your own words.)
\end{enumerate}

\textbf{Theme B: Examples of Strong and Weak Ideas}

\begin{enumerate}[label=\textbf{Q\arabic*.}, resume]
\item \textbf{Example of a Strong Idea} \\
Could you recall a strong research idea a student proposed recently? What made it strong?

\item \textbf{Example of a Weak Idea} \\
Could you recall a weak idea you have encountered? What issues made it weak?
\end{enumerate}

\textbf{Theme C: Use of Prior Scholarly Experience}

\begin{enumerate}[label=\textbf{Q\arabic*.}, resume]
\item \textbf{Experience Used During Evaluation} \\
(a) When assessing a new idea, what past knowledge or experiences do you typically rely on? \\
(b) Could you give one or two concrete examples of such intuitive or experience-based judgments?  
(For instance, empirical patterns such as ``Method X often fails under condition Y.'')
\end{enumerate}

\textbf{Theme D: Designing an AI Evaluator}

\begin{enumerate}[label=\textbf{Q\arabic*.}, resume]
\item \textbf{Capabilities Desired in an AI Evaluator} \\
If you were to design an AI system that evaluates research ideas, what capabilities should it have in order to resemble human reasoning?
\end{enumerate}

\subsection{Findings from the Pilot Study}

The interviews reveal two central findings that directly inform the design of DeepInstructor.

\paragraph{Key Evaluation Criteria.}
Across participants, three criteria consistently emerged as the primary factors instructors use when evaluating student ideas: \textit{Novelty}, \textit{Significance}, and \textit{Feasibility}. 
These criteria were mentioned spontaneously in response to Q1, and were further reinforced through examples in Q2 and Q3, where strong ideas were typically described as addressing a meaningful problem with a novel perspective, while weak ideas commonly failed due to feasibility issues such as unrealistic assumptions, unclear execution paths, or known limitations of the proposed techniques. 
This triangulation establishes these three criteria as the core dimensions that a human-aligned evaluator should prioritize.

\paragraph{Reliance on Scholarly Experience.}
Interviews also revealed that instructors draw heavily on accumulated scholarly experience when forming judgments. 
Through Q4, faculty frequently described relying on empirical heuristics, known failure modes of algorithms, benchmark-specific intuitions, and domain-specific expectations. These descriptions can be naturally expressed as stance-bearing statements such as 
``Method X usually fails when data are sparse'' or 
``Approach Y tends to work well only when Z holds.'' 
These stance-like judgments provide structured, reusable forms of research experience, motivating our conceptualization of academic experience as a graph of stance-bearing statements.

\paragraph{Desired Capabilities of an AI Evaluator.}
In Q5, participants emphasized that an AI evaluator should be able to:  
(i) recognize novelty within the broader literature,  
(ii) identify feasibility risks based on method limitations and empirical regularities, and  
(iii) reason using past scholarly experience rather than relying solely on parametric LLM knowledge.  
These observations directly inform the design of DeepInstructor, which evaluates ideas using dimension-specific strategies grounded in experience retrieval.

Overall, the pilot study demonstrates that human instructors rely on a small set of core criteria and rich, stance-bearing scholarly experience when evaluating research ideas. These findings motivate the Experience Graph and the experience-guided evaluation paradigm central to DeepInstructor.

\section{DeepInstruct Dataset Construction Details}
\label{appendix:dataset}

\begin{figure}
    \includegraphics[width=\linewidth]{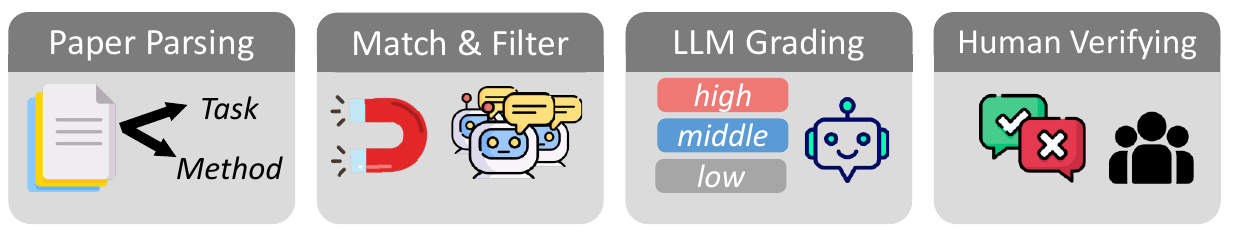}
    \caption{Pipeline of DeepInstruct Dataset Construction.}
    \label{dataset}
\end{figure}

To date, the evaluation of early-stage research ideas remains constrained by a lack of dedicated datasets \citep{xie2025far}. Directly soliciting feedback from domain experts is highly resource-intensive \citep{canllm}, and present peer review data collection methods will expose experiment results, rendering them unsuitable for assessing \citep{deepreview}. To address this gap, we introduce DeepInstruct, a novel dataset comprising early-stage research ideas coupled with multi-dimensional evaluation scores. Extracted from ICLR 2026 peer reviews, DeepInstruct demonstrates strong alignment with human expert judgments. The construction pipeline is shown in Figure~\ref{dataset}.

\subsection{Extracting and Indexing Paper Tasks and Methods}

Our dataset construction began with the collection of peer reviews from 19,398 ICLR papers after removing submissions lacking review data. This provides an abundant source for evaluation. The massive scale of the data enables us to trade recall for higher precision. Then we employed GPT-4o-mini to systematically extract the core task and proposed methodology of each paper. Recognizing the nuanced distinctions among various research tasks, we designed our prompts to elicit comprehensive problem descriptions that integrate both contextual background and specific objectives. To maintain clear conceptual boundaries, the model was explicitly constrained from generating overlapping content between the task and method sections. Subsequently, considering the brevity of the task and method descriptions, we leveraged text-embedding-v4 to obtain dense representations of the titles and methods. This embedding step streamlines the downstream organization, allowing us to effectively cluster topically and methodologically similar papers into a candidate pool easily.

\subsection{Organizing Papers with the Same Task and Method}

\textbf{Embedding Similarity Search.} Following the previous stage, our initial set of papers $\mathbf{P} = \{P_i\}_{i=1}^N$ is parsed into a corresponding set of task-method pairs $\mathbf{TM} = \{(T_i, M_i)\}_{i=1}^N$, where $N = 19,398$. To comprehensively group papers sharing either identical tasks or identical methods, we conduct two parallel retrieval processes. 

First, we randomly sample an index subset $\mathcal{Q}^T \subset \{1, 2, \dots, N\}$ with a size of $|\mathcal{Q}^T| = 1,000$ to serve as \textit{task queries}. For each query index $q \in \mathcal{Q}^T$, we utilize its task embedding vector, denoted as $\mathbf{v}_{q}^T$, to perform semantic matching against all tasks in $\mathbf{TM}$, retrieving the index set $\mathcal{R}_q^T$ of the top-10 most similar papers ($|\mathcal{R}_q^T| = 10$). 

Similarly, we randomly sample another index subset $\mathcal{Q}^M \subset \{1, 2, \dots, N\}$ of size $|\mathcal{Q}^M| = 1,000$ as \textit{method queries}. Using the method embedding $\mathbf{v}_{q}^M$ for each $q \in \mathcal{Q}^M$, we retrieve the top-10 index set $\mathcal{R}_q^M$ by semantically matching against all methods. This top-10 threshold is chosen empirically; our observations indicate that papers ranked beyond this cutoff rarely share the exact same task or method.

Consequently, for the task-driven retrieval, we form candidate clusters $C_q^T = \{(T_k, M_k) \mid k \in \mathcal{R}_q^T\}$. For the method-driven retrieval, we form clusters $C_q^M = \{(T_k, M_k) \mid k \in \mathcal{R}_q^M\}$. The final output of this stage comprises two structured collections of candidate pools: the task-centric clusters $\mathbf{C}^T = \{C_q^T \mid q \in \mathcal{Q}^T\}$ and the method-centric clusters $\mathbf{C}^M = \{C_q^M \mid q \in \mathcal{Q}^M\}$.

\textbf{Selected LLMs as a Filter.} Following the embedding-based clustering, we introduce Large Language Models as an advanced filter to rigorously filter out papers that share the exact same tasks or methods. Taking task filtering as an example, we propose an evaluation pipeline comprising human annotation probing and LLM combination for alignment. First, domain experts are instructed to annotate a subset of 150 candidate clusters, manually selecting the papers that share the same task as the query. Subsequently, we evaluate the consistency between the LLMs' predictions and the human annotations. We employed multiple LLMs and experimented with various combination strategies using set intersection and union. Ultimately, we adopted the LLM combination that achieved the highest precision to conduct the final filtering. The alignment results are presented in Table \ref{tab:model_combinations}.

\begin{table*}[htbp]
    \centering
    \caption{Performance of Different Model Combinations (Top 6 by Precision)}
    \label{tab:model_combinations}
    \resizebox{\textwidth}{!}{ 
    \begin{tabular}{llccc}
        \toprule
        \textbf{Combination} & \textbf{Type} & \textbf{Precision} & \textbf{Recall} & \textbf{F1} \\
        \midrule
        gpt4\_turbo + gpt4o + qwen\_plus & intersection & 0.7824 & 0.3788 & 0.3948 \\
        gemini + gpt4\_turbo + qwen\_plus & intersection & 0.7744 & 0.3809 & 0.3913 \\
        gpt4\_turbo + qwen\_plus & intersection & 0.7634 & 0.4000 & 0.4140 \\
        gpt4\_turbo + gpt4o\_mini + qwen\_plus & intersection & 0.7631 & 0.3675 & 0.3819 \\
        gemini + gpt4\_turbo + gpt4o & intersection & 0.7554 & 0.3654 & 0.3696 \\
        gpt4\_turbo + gpt4o & intersection & 0.7297 & 0.3896 & 0.3981 \\
        \midrule
    \end{tabular}
    }
\end{table*}

\subsection{LLM Grading}

For each filtered cluster sharing an identical task, we employed gpt-4o-mini to synthesize the corresponding peer reviews and assign categorical grades (High, Medium, or Low) across two dimensions: novelty and feasibility. Conversely, for clusters sharing an identical method, the model was tasked with evaluating the significance dimension. This design strictly adheres to our evaluation taxonomy, which posits that significance inherently reflects the value of the target task, whereas novelty and feasibility assess the design of the proposed methodology. To ensure the reliability and objectivity of the grading process, we utilized a Chain-of-Thought (CoT) prompting strategy. The LLM was explicitly instructed to articulate its reasoning based solely on the provided expert reviews before outputting a final grade, thereby preventing any subjective self-judgment or injection of external bias. They were prompted refuse to answer if there is not any information mentioned in the reviews. Furthermore, the models were explicitly prompted to decline evaluation if the required information was absent from the provided reviews.
Finally, within each cluster and across each evaluation dimension, we construct pairwise comparisons between papers that received distinct grades.

\subsection{Human Evaluating}
After completing the above steps, human experts verified each piece of data one by one. The human experts were required to check whether the tasks or methods were consistent and whether they agreed with the level of automatic annotation by AI.
Finally, we convert the paper into the idea format, that is, to hide the experimental results and focus on describing the problem, method, and experimental design. The DeepInstruct dataset comprises 398 ideas, containing a total of 289 pairwise comparison records. Broken down by evaluation dimension, feasibility includes 130 pairs involving 212 ideas; novelty consists of 129 pairs involving 206 ideas; and significance accounts for 30 pairs involving 56 ideas.

\section{Methodology Details}
\subsection{Performance of LLM scorer}
\label{sec:scorer}

\begin{table}[htbp] 
\centering
\setlength{\tabcolsep}{4pt} 
\caption{Comparison of Novelty score distributions. Scores from 1 to 5 are omitted (counts are zero).}
\label{tab:score_distribution}
\begin{tabular}{lccccc}
\toprule
\multirow{2}{*}{\textbf{Method}} & \multicolumn{5}{c}{\textbf{Score (\%)}} \\ 
\cmidrule(lr){2-6}
 & \textbf{6} & \textbf{7} & \textbf{8} & \textbf{9} & \textbf{10} \\
\midrule
Direct Scoring & 0.0 & 0.0 & 76.3 & 23.7 & 0.0 \\
Evaluation+Scorer & 1.5 & 18.5 & 3.0 & 43.0 & 34.1 \\
\bottomrule
\end{tabular}
\end{table}

The proposed LLM Scorer demonstrated strong predictive capability when estimating actual reviewer scores based solely on textual reviews. The scorer can accurately gauge the degree of polarity in the evaluation, resulting in a significantly more uniform distribution of scores (as shown in Table~\ref{tab:score_distribution}). The predicted scores achieved an average accuracy of 75.80\% within ±1.0 point and 93.87\% within ±2.0 points of the ground-truth scores, as shown in Table~\ref{tab:task_metric_comparison}.

\begin{table*}[htbp]
\centering
\caption{Performance of LLM scorer across different evaluation dimensions.}
\label{tab:task_metric_comparison}
\renewcommand{\arraystretch}{1.3} 
\begin{tabular}{l l l l l}
\hline
\textbf{Dimension} & \textbf{Precision} & \textbf{Pearson} & \textbf{Spearman} & \textbf{Kappa} \\
\hline
Significance & $\pm 1$: 0.7232; $\pm 2$: 0.9345 & 0.8220 & 0.8473 & 0.7760 \\
Novelty & $\pm 1$: 0.6528; $\pm 2$: 0.9021 & 0.8086 & 0.8128 & 0.7551 \\
Feasibility\_score & $\pm 1$: 0.8980; $\pm 2$: 0.9796 & 0.7714 & 0.7698 & 0.6556 \\
\hline
\end{tabular}
\end{table*}

\subsection{Graph Construction Details}
\label{triplets}
The resulting Experience Graph contains 416,330 evaluative triples extracted from
58,607 peer reviews, covering 567,764 unique entities and 153,391 unique relations. On average, each review contributes 7.55
triples. These statistics show that the graph is not a sparse symbolic resource, but
a large-scale external memory of reviewer experience that captures diverse
stance-bearing judgments over tasks, methods, assumptions, and evaluation protocols.
\begin{table}[H]
\centering
\small
\begin{tabular}{lc}
\toprule
\textbf{Metric} & \textbf{Score} \\
\midrule
Problem coverage & 92.8\% \\
Method coverage & 88.6\% \\
Overall coverage & 90.7\% \\
Macro-average F1 & 0.9064 \\
\bottomrule
\end{tabular}
\caption{Coverage of reviewer-described ideas by the Experience Graph.}
\label{tab:graph_coverage}
\end{table}
To evaluate whether the Experience Graph covers the conceptual space described in peer reviews, we compare two entity sets for each review: one extracted from the paper's problem and method descriptions, and the other extracted from the review-derived graph triples. Two entities are considered matched if their BGE-M3 embedding similarity exceeds 0.7. As shown in Table~\ref{tab:graph_coverage}, the
graph achieves 92.8\% problem coverage, 88.6\% method coverage, 90.7\% overall
coverage, and a macro-average F1 of 0.9064. These results suggest that although the graph is constructed only from ICLR 2024 and 2025 reviews, it captures the majority of evaluative concepts appearing in reviewer-described research ideas.

\begin{figure*}[htbp]
\small

\begin{tcolorbox}[enhanced,colback=purple!2,colframe=purple!60!blue,colbacktitle=purple!60!blue,coltitle=white,title={\textbf{Prompt of Triplet Extraction}},fonttitle=\small,boxrule=0.6pt,left=2mm,right=2mm,top=1mm,bottom=1mm]

You are a professional academic evaluation experience extraction expert. 
Please extract detailed and specific knowledge entities and experiential relationships from the following review text to construct an experiential relationship subgraph. 

Knowledge entities include but are not limited to questions, methods, concepts, theories, scenarios, and other professional terms and knowledge. Please do not use a simple word; it is better to enrich the semantic meaning of the entity based on the original evaluation by adding some adjectives. 

Experience relations refer to, for example, ``seems to be relatively good at improving... ability in xxx'', ``seems unable to be well achieved through... '' and other similar evaluative relationship statements. Their characteristics are: they have certain positive or negative emotional evaluation information, and they have as detailed as possible semantic information of specific aspects. Nodes and edges must directly depend on the review. The original text is only for reference. Therefore, when each edge is constructed, there must be corresponding evidence.

\vspace{1mm}
\noindent\textbf{Entity Naming Guidelines - Be Specific and Contextual:}

\noindent\textcolor{red}{\textbf{[X]}} \textbf{WRONG Examples (too generic):}
\begin{itemize}[leftmargin=*, nosep]
    \item ``LLMs'' $\rightarrow$ (It depends on the original text specifically.) Might be ``scientific reasoning ability of current large language models''
    \item ``performance'' $\rightarrow$ Might be ``few-shot learning ability of current large language models'' 
    \item ``dataset'' $\rightarrow$ Might be ``SCIBENCH scientific reasoning evaluation dataset''
\end{itemize}

\vspace{1mm}
\noindent\textcolor{green!60!black}{\textbf{[V]}} \textbf{CORRECT Examples (specific and contextual):}
\begin{itemize}[leftmargin=*, nosep]
    \item ``scientific reasoning ability of current large language models''
    \item ``few-shot learning performance on complex scientific problems''
    \item ``SCIBENCH scientific reasoning evaluation dataset''
    \item ``Chain-of-Thought (CoT) prompting technique''
\end{itemize}

\vspace{1mm}
\noindent\textbf{Relationship Examples - Be Evaluative and Specific:}

\noindent\textcolor{green!60!black}{\textbf{[V]}} \textbf{Good Relationship Examples:} (Include descriptions of both positive and negative aspects and degrees, as well as rich semantic information)
\begin{itemize}[leftmargin=*, nosep]
    \item ``Chain-of-Thought (CoT) prompting'' $\rightarrow$ ``significantly improves'' $\rightarrow$ ``calculation skills of LLMs''
    \item ``SCIBENCH dataset'' $\rightarrow$ ``effectively differentiates'' $\rightarrow$ ``performance between different large language models''
    \item ``free-form questions'' $\rightarrow$ ``a certain degree prevents'' $\rightarrow$ ``result guessing based on multiple-choice answers''
    \item ``systematic zero-shot example selection'' $\rightarrow$ ``can not enhances'' $\rightarrow$ ``model's problem-solving capabilities''
    \item ``Wolfram Language prompts'' $\rightarrow$ ``deteriorates'' $\rightarrow$ ``few-shot learning performance of LLMs''
\end{itemize}

\vspace{1mm}
\noindent\textcolor{red}{\textbf{[X]}} \textbf{WRONG Relationship Examples (too generic):}
\begin{itemize}[leftmargin=*, nosep]
    \item ``LLMs'' $\rightarrow$ ``is designed to'' $\rightarrow$ ``performance'' (too vague)
    \item ``Error analysis'' $\rightarrow$ ``requires understanding of'' $\rightarrow$ ``Hückel molecular orbital theory'' (knowledge relation, not evaluative)
\end{itemize}

\vspace{1mm}
\noindent\textbf{Attention}
\begin{itemize}[leftmargin=*, nosep]
    \item[1.] The evidence section must be complete and no part should be omitted. Never use ``...''.
    \item[2.] Node and Edge should be as specific and detailed as possible, and when combined, they should conform to the logic of the original text.
\end{itemize}

\vspace{1mm}
\noindent You should check the review sentence by sentence to see if there are any evaluations from the reviewers regarding the method or professional terms, and then extract them. And speak out your thinking process aloud. 
Your final output should be in JSON format:

\vspace{1mm}
\noindent\texttt{\{\{} \\
\texttt{\ \ \ \ "source\_name": "node\_x",} \\
\texttt{\ \ \ \ "target\_name": "node\_y",} \\
\texttt{\ \ \ \ "relationship": "xxx",} \\
\texttt{\ \ \ \ "evidence": "relevant sentence from review text (Or the rewritten text in the Node process result)"} \\
\texttt{\}\}}

\vspace{1mm}
\noindent\textbf{Example input:} The proposed method cannot demonstrate its effectiveness on both image and audio datasets. And BATTLE can improve the robustness of training agents under adversarial attacks.

\vspace{1mm}
\noindent\textbf{Sample json output} (omitting the previous thinking process):

\noindent\texttt{[\{\{} \\
\texttt{\ \ \ \ "source\_name": "PIA",} \\
\texttt{\ \ \ \ "target\_name": "some image and audio datasets",} \\
\texttt{\ \ \ \ "relationship": "is NOT validated on",} \\
\texttt{\ \ \ \ "evidence": "The proposed method cannot demonstrate its effectiveness on both image and audio datasets. PS: Based on the paper, the proposed method refer to PIA."} \\
\texttt{\}\},} \\
\texttt{\{\{} \\
\texttt{\ \ \ \ "source\_name": "BATTLE",} \\
\texttt{\ \ \ \ "target\_name": "adversarial attacks of agents",} \\
\texttt{\ \ \ \ "relationship": "enhance the robustness of",} \\
\texttt{\ \ \ \ "evidence": "And BATTLE can improve the robustness of training agents under adversarial attacks."} \\
\texttt{\}\}]}

\vspace{1mm}
Be specific and detailed. Now the review text is: \texttt{\{review\_text\}}
And the pdf url is (I'll also input the pdf to you to get the original text): \\
\texttt{\{pdf\_url\}}
\end{tcolorbox}
\vspace{-2mm}
\caption{System prompt utilized for Experience Graph Extraction.}
\label{fig:extraction_prompt}
\end{figure*}

\section{Evaluation Metrics}
\label{deepintrcut:metrics}

This section provides the detailed definitions and implementation protocol for the metrics used in Section~4.2.

\paragraph{Score Alignment.}
For \textsc{ExpertEval}, we compute Hit@$\pm k$ ($k \in \{1,2\}$), defined as the proportion of predictions that fall within a tolerance range of the ground-truth human scores: $\mathrm{HitRate}_{\pm k} = \frac{1}{N} \sum_{i=1}^{N} \mathbb{I}\!\left( |s_i - g_i| \le k \right),$
where $s_i$ and $g_i$ denote the predicted and ground-truth scores for the $i$-th idea, respectively. This metric is motivated by inter- rater variance in human evaluation, where small deviations often indicate practical agreement rather than disagreement.

\paragraph{Concern-Level Alignment.}
DeepInstructor's feasibility output consists of complex concerns that cannot be adequately captured by a single scalar score. We therefore design a concern-matching metric to evaluate fine-grained alignment between generated feasibility concerns and human annotations.

Given the set of human-annotated concerns $\mathcal{H}_i$ and model-generated concerns $\mathcal{M}_i$ for idea $i$, we use an LLM-based evaluator to determine whether each generated concern semantically matches any human concern. We adopt \textit{DeepSeek--Reasoner} as the evaluator because the matching process requires explicit reasoning over scientific concerns rather than surface-form overlap.

We compute two metrics: \textit{coverage} measures the proportion of human concerns identified by the model, while \textit{precision} measures the proportion of model-generated concerns that are valid. We report the mean coverage and mean precision across all evaluated ideas.

\section{Experiment Details}
\subsection{Detailed Results of AI in the Loop}
\label{loop}

Table~\ref{tab:feedback_scores} shows detailed AI in the loop experiment results.

\begin{table*}[htbp] 
\centering
\caption{Feedback scores and relative changes (compared to $R_0$) across different dimensions and iterations.}
\label{tab:feedback_scores} 
\resizebox{\textwidth}{!}{ 
\begin{tabular}{llccccc}
\toprule
\textbf{Type} & \textbf{Dimension} & \textbf{$R_0$} & \textbf{$R_1$} & \textbf{$R_2$} & \textbf{$R_3$} & \textbf{Avg. Change} \\
\midrule
\multirow{3}{*}{\makecell[l]{ChatGPT-4o \\ Instructor feedback}} 
 & Feasibility  & 6.203 & 6.359 \textcolor{blue!60}{\footnotesize (+0.156)} & 6.224 \textcolor{blue!60}{\footnotesize (+0.021)} & 6.218 \textcolor{blue!60}{\footnotesize (+0.015)} & +0.064 \\
 & Novelty      & 5.802 & 5.880 \textcolor{blue!60}{\footnotesize (+0.078)} & 6.191 \textcolor{blue!60}{\footnotesize (+0.389)} & 6.202 \textcolor{blue!60}{\footnotesize (+0.400)} & +0.289 \\
 & Significance & 8.270 & 8.509 \textcolor{blue!60}{\footnotesize (+0.239)} & 8.355 \textcolor{blue!60}{\footnotesize (+0.085)} & 8.716 \textcolor{blue!60}{\footnotesize (+0.446)} & +0.257 \\
\midrule
\multirow{3}{*}{\makecell[l]{ChatGPT-4o-mini \\ Instructor feedback}} 
 & Feasibility  & 5.569 & 5.559 \textcolor{red!60}{\footnotesize (-0.010)} & 5.582 \textcolor{blue!60}{\footnotesize (+0.013)} & 5.639 \textcolor{blue!60}{\footnotesize (+0.070)} & +0.024 \\
 & Novelty      & 5.914 & 6.022 \textcolor{blue!60}{\footnotesize (+0.108)} & 6.125 \textcolor{blue!60}{\footnotesize (+0.211)} & 5.922 \textcolor{blue!60}{\footnotesize (+0.008)} & +0.109 \\
 & Significance & 8.612 & 8.752 \textcolor{blue!60}{\footnotesize (+0.140)} & 8.899 \textcolor{blue!60}{\footnotesize (+0.287)} & 8.876 \textcolor{blue!60}{\footnotesize (+0.264)} & +0.230 \\
\midrule
\multirow{3}{*}{\makecell[l]{ChatGPT-4o-mini \\ prompt feedback}} 
 & Feasibility  & 5.569 & 5.640 \textcolor{blue!60}{\footnotesize (+0.071)} & 5.678 \textcolor{blue!60}{\footnotesize (+0.109)} & 5.594 \textcolor{blue!60}{\footnotesize (+0.025)} & +0.068 \\
 & Novelty      & 5.914 & 5.884 \textcolor{red!60}{\footnotesize (-0.030)} & 5.845 \textcolor{red!60}{\footnotesize (-0.069)} & 6.039 \textcolor{blue!60}{\footnotesize (+0.125)} & +0.009 \\
 & Significance & 8.612 & 8.791 \textcolor{blue!60}{\footnotesize (+0.179)} & 8.736 \textcolor{blue!60}{\footnotesize (+0.124)} & 8.693 \textcolor{blue!60}{\footnotesize (+0.081)} & +0.128 \\
\bottomrule
\end{tabular}
}
\end{table*}

\subsection{Case Study}
\label{case-sd}

\begin{figure*}[htbp]
    \centering
    \includegraphics[width=\textwidth]{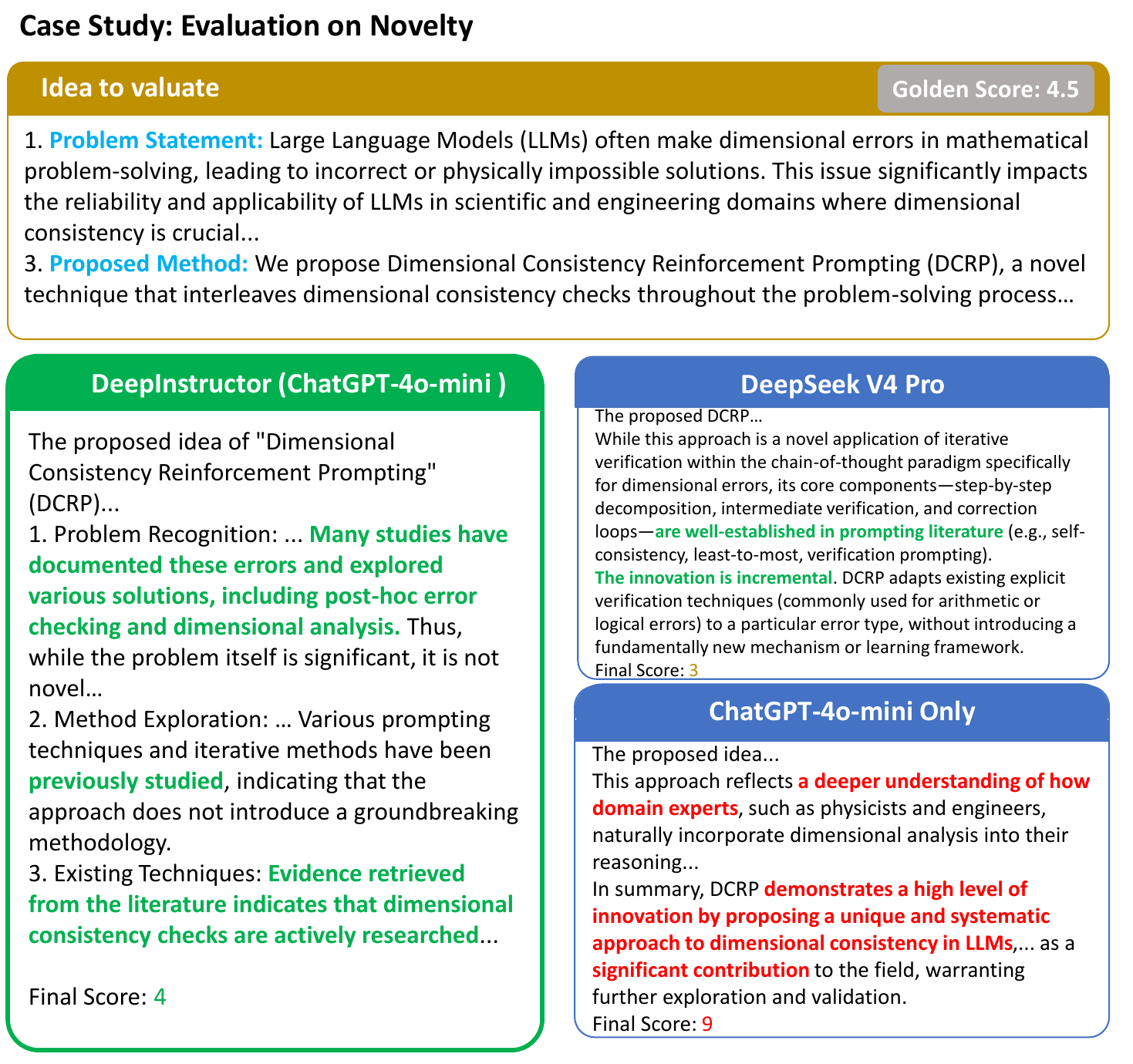}
    \caption{Case study on novelty evaluation. }
    \label{case_nov}
\end{figure*}

\begin{figure*}[htbp]
    \centering
    \includegraphics[width=\textwidth]{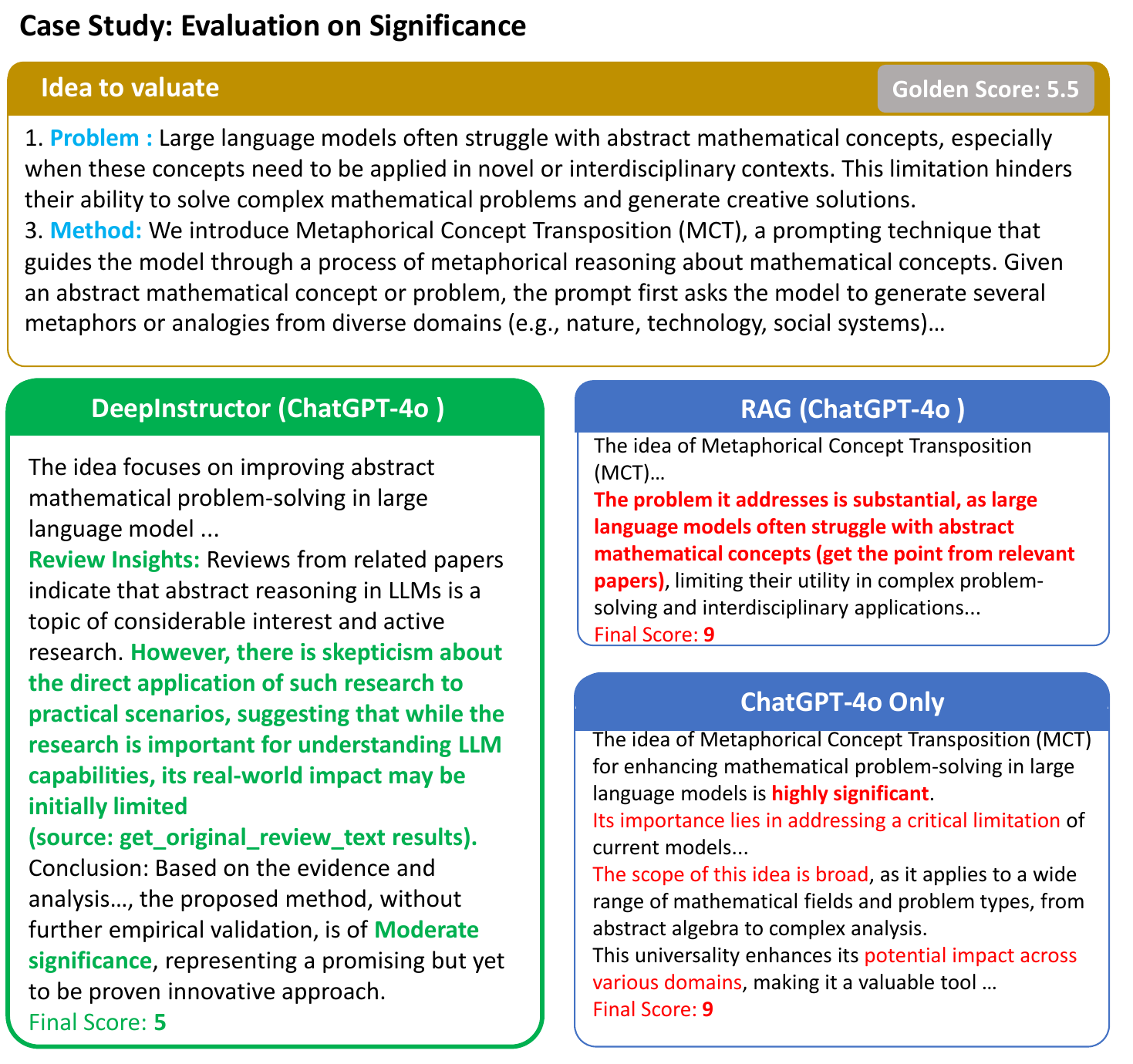}
    \caption{Case study on significance evaluation. }
    \label{case_sig}
\end{figure*}

To explain the experiment results and directly understand the pipeline of DeepInstructor, we select 2 cases as shown in Figure~\ref{case_nov} and \ref{case_sig}. Figure~\ref{fig:case_full} shows a full reasoning process of DeepInstructor. 

\paragraph{Case Study on Novelty Evaluation.}
\label{case_study_novel}
Figure \ref{case_nov} illustrates why DeepInstructor and DeepSeek-V4-Pro outperforms ChatGPT-4o-mini in evaluating novelty. As a newer model, DeepSeek-V4-Pro has ingested a vast corpus of contemporary academic literature during its pretraining. During the reasoning process, this robust parametric knowledge enables the model to effectively recall relevant prior work, thereby facilitating a highly accurate novelty assessment. In contrast, ChatGPT-4o-mini is constrained by an earlier knowledge cutoff and lacks exposure to recent publications, such as ICLR 2024 papers. This knowledge deficit restricts its evaluative capability, leading to a severe positive bias where it consistently overestimates novelty and assigns inflated scores.

DeepInstructor can access both positive and negative evaluations of relevant knowledge entities, allowing it to provide a precise final evaluation, unlike DeepSeek's black-box and overcritical approach.

\paragraph{Case Study on Significance Evaluation.}
\label{case_study_sig}
Figure~\ref{case_sig} has shown how DeepInstructor outperforms pure ChatGPT-4o model and RAG method when evaluating significance.
In this example, DeepInstructor retrieves reviewer experience related to abstract reasoning and metaphorical reasoning in LLMs, and further identifies skeptical reviewer judgments regarding their direct applicability to practical scenarios. Grounded in such retrieved experience, DeepInstructor produces a more conservative significance assessment.

In contrast, both RAG and standalone LLM evaluation tend to assign overly positive significance judgments. Although RAG retrieves semantically relevant content, it fails to retrieve sufficiently critical evaluative evidence, while standalone LLMs rely primarily on parametric knowledge. As a result, both systems overestimate the real-world significance of the proposed idea and produce substantially inflated scores.

We further observe that significance evaluation remains considerably more challenging than novelty and feasibility evaluation across all methods. Manual inspection suggests that LLMs exhibit an inherent tendency to overestimate significance, particularly for ideas framed around broad or impactful research problems. Moreover, peer reviews themselves contain relatively sparse explicit discussion regarding significance compared to methodological novelty or feasibility concerns, limiting the amount of directly retrievable evaluative experience.

\begin{figure*}[htbp]
\small

\begin{tcolorbox}[enhanced,colback=purple!2,colframe=purple!60!blue,colbacktitle=purple!60!blue,coltitle=white,title={\textbf{Novelty Evaluation: Chain-of-Quote Prompting Improves Factuality and Attribution in Multi-Hop Reasoning}},fonttitle=\small,boxrule=0.6pt,left=2mm,right=2mm,top=1mm,bottom=1mm]
\centerline{\small \textbf{Golden Score:} 1 4 3\quad\textbf{DeepInstructor Score:} 2.0\quad\textbf{Pure DeepSeek Score:} 7.0}

\textbf{Step 1:} Let me search for the key concepts in idea. \\

\textbf{Step 2:} \textcolor{green!40!black}{\texttt{\footnotesize search(node="LLM quoting from pretraining data", edge="has been researched widely")}} \\

\textbf{Step 3:}
\quad $\bullet$ \textcolor{green!40!black}{\texttt{\scriptsize multi-agent collaborative data selection mechanism}} $\rightarrow$ \textcolor{green!30!black}{\textit{\scriptsize significantly improves}} $\rightarrow$ \textcolor{green!40!black}{\texttt{\scriptsize data efficiency during LLM pretraining}} \\
\quad $\bullet$ \textcolor{green!40!black}{\texttt{\scriptsize Multi-Agent Collaborative Data Selection Framework}} $\rightarrow$ \textcolor{green!30!black}{\textit{\scriptsize significantly enhances}} $\rightarrow$ \textcolor{green!40!black}{\texttt{\scriptsize data efficiency of LLM pretraining}} \\
\quad $\bullet$ \textcolor{green!40!black}{\texttt{\scriptsize query rewriting}} $\rightarrow$ \textcolor{green!30!black}{\textit{\scriptsize may introduce risks of hallucination}} $\rightarrow$ \textcolor{green!40!black}{\texttt{\scriptsize knowledge in LLM’s pretraining data}} \\
\quad $\bullet$ \textcolor{green!40!black}{\texttt{\scriptsize Task-gram Language Model}} $\rightarrow$ \textcolor{green!30!black}{\textit{\scriptsize significantly captures}} $\rightarrow$ \textcolor{green!40!black}{\texttt{\scriptsize relationship between LLM performance and pretraining data}} \\
\quad $\bullet$ \textcolor{green!40!black}{\texttt{\scriptsize pre-training data of the LLM}} $\rightarrow$ \textcolor{green!30!black}{\textit{\scriptsize limits}} $\rightarrow$ \textcolor{green!40!black}{\texttt{\scriptsize novelty in generated problems}} \\
SUM: Based on the results, [LLM quoting from pretraining data] is identified as a PROBLEM. The assessment will focus solely on the problem's innovation level. The retrieved evidence indicates that the relationship between LLM performance and pretraining data has been a subject of research, with specific references to limitations in novelty.\\

\noindent\textbf{Step 4:} Ok. Let me take a look at the concept of the method.\\

\textbf{Step 5:} \textcolor{green!40!black}{\texttt{\footnotesize search(node="Chain-of-Verification prompting", edge="is a method to solve")}} \\

(... 30 steps omitted)

\noindent\textbf{Final Step:} {\color{red!60!black}\small I now have enough evidence to make a thorough evaluation. Let me compile my analysis.

\noindent\textbf{\normalsize Innovation Evaluation: "Chain-of-Quote Prompting Improves Factuality and Attribution in Multi-Hop Reasoning"}

\noindent\textbf{\small Analysis of the Idea's Components}

\textbf{Core problem addressed:} LLM hallucination and lack of attribution in multi-hop reasoning.

\textbf{Core method proposed:} A prompting technique that (1) instructs the model to reason step-by-step, (2) at each step, decides whether to quote from a knowledge corpus (e.g., Wikipedia), and (3) when quoting is chosen, generates multiple candidates and selects the one with the highest QUIP-Score (a measure of verbatim quoting from pretraining data).

\noindent\textbf{\small Evidence Synthesis}

\textbf{1. The problem space is well-trodden.} The evidence consistently shows that:
\quad $\bullet$ LLM hallucination in multi-hop reasoning is a widely recognized problem with numerous existing solutions (RAG, Chain-of-Knowledge, Staple, code-writing and grounding techniques, Mask-DPO, etc.)
\quad $\bullet$ Attribution in LLM reasoning is also an actively researched area with established methods (TRACE framework, watermark-based attribution, etc.)
\quad $\bullet$ Multi-hop reasoning itself has been extensively studied with many proposed methods (CoT, CoVe, self-verification, etc.)

\textbf{2. The individual components of the method are not novel:}
\quad $\bullet$ \textbf{Step-by-step reasoning (Chain-of-Thought):} This is a well-established technique with thousands of papers. The evidence explicitly notes that CoT is a "widely used technique" and has "moderate" innovation.
\quad $\bullet$ \textbf{Generating multiple candidates and selecting the best one:} This is a common paradigm in LLM reasoning (self-consistency, SelfCheck, ToG, pairwise comparisons). The evidence confirms this approach has been "partially addressed."
\quad $\bullet$ \textbf{Using QUIP-Score for measuring quoting:} The evidence found no direct references to QUIP-Score in the database, suggesting it is a niche or very recent metric, but the concept of measuring verbatim overlap with pretraining data is itself not fundamentally new.
\quad $\bullet$ \textbf{Prompting the model to decide whether to quote:} The evidence explicitly states that "the high-level idea of using prompt to instruct LLM is not new."

\textbf{3. The specific combination is the main claim of novelty.} The idea's primary innovation lies in \textit{integrating} quoting decisions into the reasoning chain, rather than treating quoting as a separate post-hoc verification step. However, this integration is incremental:
\quad $\bullet$ \textbf{Chain-of-Knowledge (CoK)} already integrates knowledge verification into reasoning chains.
\quad $\bullet$ \textbf{RAG-based methods} already retrieve and incorporate external knowledge into reasoning.
\quad $\bullet$ The "decide whether to quote" step is essentially a retrieval-triggering mechanism, which is conceptually similar to how RAG systems decide when to retrieve.

\textbf{4. The examples provided (The Police / warthog on Broadway) demonstrate a straightforward application} that does not reveal a fundamentally new reasoning capability. The model is essentially being asked to decompose a question and look up facts — a process that existing RAG+CoT pipelines already accomplish.

\noindent\textbf{\small Innovation Level Judgment: **LOW to MODERATE**}

\textbf{Reasoning:}
\quad $\bullet$ The idea does \textbf{not} introduce a new problem; it addresses the well-studied problem of hallucination and attribution in multi-hop reasoning.
\quad $\bullet$ The individual techniques (step-by-step reasoning, candidate generation, quoting-based selection) are all established in the literature.
\quad $\bullet$ The combination of these components, while perhaps not previously published in this exact configuration, represents an \textbf{incremental integration} of existing ideas rather than a conceptual breakthrough.
\quad $\bullet$ The core mechanism — using verbatim quoting from pretraining data as a signal for factuality — builds directly on prior work by Weller et al. (2024) showing that LLMs quote from pretraining data. Extending this observation from an analytical finding to a prompting technique is a natural and predictable next step.
\quad $\bullet$ The method does not introduce new theoretical insights or fundamentally change how LLMs reason; it simply adds a quoting-based selection mechanism to existing CoT pipelines.

The idea is \textbf{not entirely without merit} — the specific design of letting the model decide when to quote and using QUIP-Score for candidate selection is a thoughtful engineering contribution. However, as an \textit{innovation}, it falls short of being considered highly novel, as it largely recombines well-known techniques in a predictable manner.}
\end{tcolorbox}

\vspace{-2mm}
\caption{Case study on \textbf{Novelty}: DeepInstructor (DeepSeek). \textcolor{blue!60!black}{Blue} = Thinking, \textcolor{green!40!black}{Green} = Retrieval, \textcolor{red!60!black}{Red} = Final Evaluation.}
\label{fig:case_full}
\end{figure*}

\section{Why Hit@1/Hit@2 Are Superior to Spearman Correlation for Measuring Evaluation Consistency}
\label{sec:proof}

Formally,
\[
\mathrm{HitRate}_{\pm k} = \frac{1}{N} \sum_{i=1}^{N} \mathbb{I}\!\left( |s_i - g_i| \le k \right),
\]
where $s_i$ and $g_i$ denote the predicted and ground-truth scores, respectively, and $k \in \{1,2\}$.

\subsection{Evidence of High Variance Among Reviewers}
\label{sec:reviewer_variance}

To understand the inherent disagreement in human evaluation, we analyze 19,433 papers from ICLR 2026 with a total of 112,001 reviewer pairs. This analysis reveals that significant disagreement among reviewers is the norm rather than the exception.

\subsubsection{Distribution of Reviewer Score Differences}
\newcommand{\hit}[1]{\text{Hit@#1}}
For each paper, we compute the absolute difference between all pairs of reviewers. Table~\ref{tab:diff_distribution} presents the distribution of these differences (ICLR uses an even-numbered rating scale).

\begin{table}[htbp] 
\centering
\caption{Distribution of Absolute Differences Between Reviewer Pairs}
\label{tab:diff_distribution}
\resizebox{\columnwidth}{!}{
\begin{tabular}{lccc} 
\toprule
\textbf{Diff.} & \textbf{Count} & \textbf{\%} & \textbf{Cumul. (\%)} \\
\midrule
0 & 37,993 & 33.92 & 33.92 \\
2 & 51,888 & 46.33 & 80.25 \\
$\geq 4$ & 22,120 & 19.75 & 100.0 \\
\midrule
\textbf{Total} & \textbf{112,001} & \textbf{100.0} & -- \\
\bottomrule
\end{tabular}
}
\end{table}

\textbf{Key Findings:}
\begin{itemize}
    \item Only \textbf{33.9\%} of reviewer pairs achieve perfect agreement (difference = 0)
    \item \textbf{66.1\%} of reviewer pairs show disagreement (difference $\geq 2$)
    \item \textbf{19.7\%} of reviewer pairs have substantial disagreement (difference $\geq 4$)
\end{itemize}

\subsubsection{Paper-Level Score Variance}

Table~\ref{tab:range_distribution} shows the distribution of score ranges (max - min) within each paper.

\begin{table}[htbp]
\centering
\caption{Distribution of Score Range per Paper}
\label{tab:range_distribution}
\begin{tabular}{lcc}
\toprule
\textbf{Range Threshold} & \textbf{Count} & \textbf{Percentage} \\
\midrule
Range $\geq 2$ & 18,117 & 93.2\% \\
Range $\geq 4$ & 10,052 & 51.7\% \\
\midrule
Average Range & \multicolumn{2}{c}{$3.16 \pm 1.63$} \\
\bottomrule
\end{tabular}
\end{table}

\textbf{Implications:}
\begin{itemize}
    \item Over \textbf{93\%} of papers have reviewer score ranges $\geq 2$ points
    \item Over \textbf{50\%} of papers have reviewer score ranges $\geq 4$ points
    \item This demonstrates that human reviewers frequently disagree, making it essential to use metrics that can appropriately handle such disagreement
\end{itemize}

\begin{figure*}[htbp]
\centering
\includegraphics[width=0.95\textwidth]{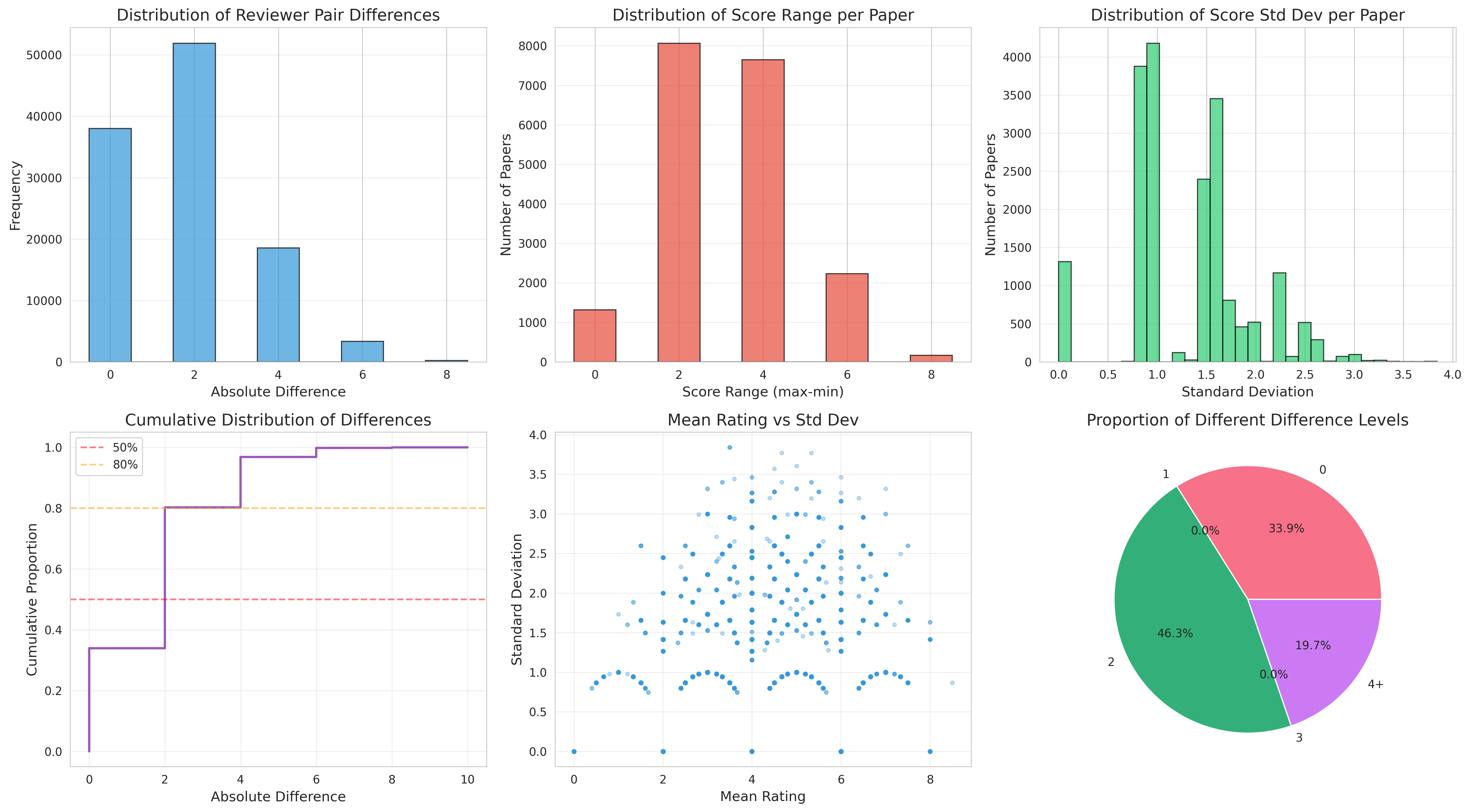}
\caption{Distribution of reviewer score differences. (a) Histogram of absolute differences. (b) Distribution of score range per paper. (c) Distribution of standard deviation. (d) Cumulative distribution. (e) Mean vs. std scatter plot. (f) Proportion pie chart.}
\label{fig:reviewer_distribution}
\end{figure*}

Figure~\ref{fig:reviewer_distribution} provides visualizations of these distributions.

\subsection{Spearman Correlation Fails to Measure Reviewer-Reviewer Consistency}
\label{sec:spearman_fails}

To demonstrate the limitations of Spearman correlation, we treat each pair of reviewers as two independent ``judges'' and compute consistency metrics between them.

\subsubsection{Experimental Method}

For each reviewer pair $(r_1, r_2)$ from the same paper:
\begin{itemize}
    \item Treat $r_1$ as ``AI prediction''
    \item Treat $r_2$ as ``human annotation''
    \item Compute $\hit{1}$, $\hit{2}$, Pearson correlation, and Spearman correlation
\end{itemize}

This setup simulates the scenario of evaluating an AI system against human judgments, but uses actual human reviewers on both sides.

\subsubsection{Overall Consistency Metrics}

Table~\ref{tab:overall_metrics} presents the consistency metrics across all 112,001 reviewer pairs.

\begin{table*}[htbp]
\centering
\caption{Consistency Metrics Across All Reviewer Pairs}
\label{tab:overall_metrics}
\begin{tabular}{lcc}
\toprule
\textbf{Metric} & \textbf{Value} & \textbf{Interpretation} \\
\midrule
$\hit{1}$ & 0.3392 & 33.92\% of pairs differ by $\le 1$ point \\
$\hit{2}$ & 0.8025 & 80.25\% of pairs differ by $\le 2$ points \\
Pearson correlation & 0.1796 & Weak positive correlation \\
Spearman correlation & 0.1759 & Weak positive correlation \\
\bottomrule
\end{tabular}
\end{table*}

\textbf{Key Observation:} While Spearman shows only weak correlation (0.18), $\hit{2}$ reveals that \textbf{80.25\%} of reviewer pairs actually agree within a reasonable range ($\pm 2$ points). And if the scoring range is continuous, the value of Hit@1 will increase.

\subsubsection{Metrics by Disagreement Level}

Table~\ref{tab:metrics_by_group} breaks down the metrics by reviewer difference groups.

\begin{table*}[htbp]
\centering
\caption{Consistency Metrics by Reviewer Difference Group}
\label{tab:metrics_by_group}
\begin{tabular}{lcccccc}
\toprule
\textbf{Group} & \textbf{Count} & \textbf{\%} & \textbf{$\hit{1}$} & \textbf{$\hit{2}$} & \textbf{Pearson} & \textbf{Spearman} \\
\midrule
diff = 0 & 37,993 & 33.9\% & 1.0000 & 1.0000 & 1.0000 & 1.0000 \\
diff = 2 & 51,888 & 46.3\% & 0.0000 & 1.0000 & 0.3324 & 0.3318 \\
diff $\geq 4$ & 22,120 & 19.7\% & 0.0000 & 0.0000 & \textcolor{red}{\textbf{-0.5981}} & \textcolor{red}{\textbf{-0.6012}} \\
\bottomrule
\end{tabular}
\end{table*}

\subsubsection{Critical Finding: Negative Correlation}

\textbf{When reviewer difference $\geq 4$, Spearman correlation becomes \textcolor{red}{-0.60} (strongly negative)!}

This is a critical finding that demonstrates the fundamental limitation of correlation coefficients:

\begin{enumerate}
    \item In the high disagreement group (diff $\geq 4$):
    \begin{itemize}
        \item $\hit{1} = 0$ and $\hit{2} = 0$ (as expected, since all pairs differ by $\geq 4$)
        \item However, Spearman correlation is \textbf{strongly negative}, which is misleading
    \end{itemize}

    \item This occurs because correlation measures the \textit{pattern of relationship}, not the \textit{magnitude of agreement}
\end{enumerate}

\subsection{Conclusion}

This analysis demonstrates that:
\begin{enumerate}
    \item Human reviewers show substantial disagreement 
    \item Spearman correlation \textbf{fails} to meaningfully capture consistency when disagreement is high (becomes negative)
    \item $\hit{2}$ provides a more realistic assessment: 80.25\% of human reviewer pairs agree within $\pm 2$ points
    \item For evaluating AI-human consistency, $\hit{1}$ and $\hit{2}$ are superior to Spearman correlation because they:
    \begin{itemize}
        \item Directly measure agreement rather than correlation
        \item Remain interpretable under all levels of disagreement. It conforms to the fractional characteristics of aggregation distribution
    \end{itemize}
\end{enumerate}

\section{Additional Experimental Details}
\label{sec: Experimental Details}

\subsection{Prompts for Idea Evaluation}

We used three models in total: GPT, RAG based LLM, and DeepInstructor. Evaluations were conducted across three dimensions: Novelty, Significance, and Feasibility. For the Feasibility dimension, the evaluation outputs have two forms (concerns and evaluative comments), resulting in a total of 12 designed prompts.
Tables~\ref{GPT-prompt},~\ref{RAG-prompt}, and~\ref{DeepInstructor-prompt} present the different prompts used by GPT, the RAG-based LLM, and DeepInstructor, respectively, across four output formats. The initialization sections of the prompts like \texttt{\{GPT’s\_prompt\_for\_novelty\}} in Tables~\ref{RAG-prompt} and~\ref{DeepInstructor-prompt} are identical to that in Table~\ref{GPT-prompt}, where curly braces are used as placeholders. In Table~\ref{RAG-prompt}, \texttt{\{context\}} represents the retrieved evidence. In Table~\ref{DeepInstructor-prompt}, the three tools utilized by DeepInstructor are defined within the agent's ReAct framework and therefore require no additional definition within the prompt itself. 
The 20 tool callings prompt was implemented specifically to ensure a fair comparison with the baselines; it does not compromise the inherent autonomy of the Agent. Actually, if we do not conduct this constraint, DeepInstructor will still utilize ~20 tool callings per reasoning process.

\subsection{Prompts for Few-shot LLM Scorers and DeepSeek–Reasoner}

We designed an LLM-based scorer to assign a quantitative score to the generated evaluations. 
For concerns along the feasibility dimension, we employed DeepSeeek-Reasoner to determine whether two concerns matched or not. The prompt   template for our LLM Scorer and DeepSeek–Reasoner is shown in the Table~\ref{scorer}.
\begin{table*}[htbp]
  \caption{Prompts used by GPT for evaluating each dimension.}
  \label{GPT-prompt}
  \centering
  \begin{tabular}{p{2.5cm} p{\dimexpr\textwidth-2.5cm-4\tabcolsep\relax}} 
    \toprule
    \textbf{Demonstration} & \textbf{Prompt} \\
    \midrule
    Novelty &  
    You are a professional evaluator focusing on the novelty of the idea.
    
I will provide you with an academic idea. Your task is to evaluate only its level of innovation.

Please focus exclusively on the novelty and originality of the idea — how new, unique, or creative it is compared to existing research or conventional approaches in the field.

Your response should:

Be concise and academic in tone.

Avoid discussing feasibility, impact, or methodology.

Provide a clear judgment on the innovation level with your serious analysis and reasoning. \\
    \hline
    Significance & 
    You are a professional evaluator focusing on the significance of the idea.
    
I will provide you with an academic idea. Your task is to evaluate only its level of significance.

Please focus exclusively on the importance, scope, and potential impact of the idea — how meaningful, influential, or valuable it would be compared to existing research or conventional approaches in the field.

Your response should:

Be concise and academic in tone.

Avoid discussing feasibility, novelty, or methodology.

Provide a clear judgment on the significance level with your serious analysis and reasoning.

\\
    \hline
    Feasibility 
    (Output concerns) & 
    You are a rigorous peer-reviewer.
    
Task: Critically evaluate the given idea/proposal and GENERATE potential 'concerns' of feasibility
(feasibility, feasibility doubts, missing evaluations).

Do NOT extract phrases from the text verbatim; instead, propose concerns based on your assessment.

Output Policy (STRICT):

- Return ONLY a JSON array of strings, starting with '[' and ending with ']'.

- Each item must be a single-line short sentence (no line breaks).

- Do NOT include any code fences, markdown, comments, labels, or extra text.

- No leading bullets, numbering, or trailing commas inside items.

- Aim for 8-12 high-quality, non-duplicative items covering: feasibility, feasibility doubts, missing evaluations.

\\
    \hline
    Feasibility 
    
    (Output evaluative comments) & 
    You are a rigorous peer-reviewer evaluating the feasibility of an academic research idea.
    
Task: Critically evaluate the given idea/proposal and write a comprehensive peer review evaluation text.

Your response should:
- Be a continuous, natural text similar to a peer review comment (like "all comments" in academic reviews)

- Discuss feasibility, implementation challenges, effectiveness, and potential issues

- Be concise and academic in tone

- Provide an evaluation covering both positive aspects and concerns

- Include your assessment of:

  * How easy or difficult it is to implement the idea
  
  * Whether the experimental setup is feasible
  
  * Whether the method is likely to work effectively
  
  * Any resource requirements or challenges
  
  * Comparison with existing approaches if relevant
  
- Write in a natural, flowing style as if you are providing feedback to the authors

    \\
    \bottomrule
  \end{tabular}
\end{table*}

\begin{table*}[htbp]
  \caption{Prompts used by RAG-based LLM for evaluating each dimension.}
  \label{RAG-prompt}
  \centering
  \begin{tabular}{p{2.5cm} p{\dimexpr\textwidth-2.5cm-4\tabcolsep\relax}}
    \toprule
    \textbf{Demonstration} & \textbf{Prompt} \\
    \midrule
    Novelty &  
    \{GPT's\_prompt\_for\_novelty\}
    
    Below are evidence sentences retrieved from reviews based on your query:
    \{context\}

    Please generate a novelty evaluation based on these evidence sentences and your professional knowledge for the following idea.

    Output Policy (STRICT):
    
    - Return a text evaluating innovation (which should include reasonable reasons) based on the evidence sentences.
    
    - Please evaluate the innovativeness of the idea clearly and emphatically.
    \\
    \hline
    Significance & 
    \{GPT's\_prompt\_for\_significance\}
    
    Below are evidence sentences retrieved from reviews based on your query:
    \{context\}

    Please generate a significance evaluation based on these evidence sentences and your professional knowledge for the following idea.

    Output Policy (STRICT):
    
    - Return a text evaluating **significance** (which should include reasonable reasons) based on the evidence sentences.
    
    - Please evaluate the **significance** of the idea clearly and emphatically.
\\
    \hline
    Feasibility 
    (Output concerns) & 

    \{GPT's\_prompt\_for\_feasibility\_concerns\}
    
    Below are evidence sentences retrieved from reviews based on your query:\{context\}
    
    Please generate concerns based on these evidence sentences and your professional knowledge for the following idea.

\\
    \hline
    Feasibility 
    
    (Output evaluative comments) & 
    \{GPT's\_prompt\_for\_feasibility\_comments\}
    
   Below are evidence sentences retrieved from reviews based on your query:\{context\}
    
    Please generate comments based on these evidence sentences and your professional knowledge for the following idea.

    \\
    \bottomrule
  \end{tabular}
\end{table*}

\begin{table*}[htbp]
  \caption{Prompts used by DeepInstructor for evaluating each dimension.}
  \label{DeepInstructor-prompt}
  \centering
  \begin{tabular}{p{2.5cm} p{\dimexpr\textwidth-2.5cm-4\tabcolsep\relax}}
    \toprule
    \textbf{Demonstration} & \textbf{Prompt} \\
    \midrule
    Novelty &  
    \{GPT's\_prompt\_for\_novelty\}
    
    You can use tools to learn about relevant experiences related to innovation. Core argument explanation:

    node\_query: Enter a specific knowledge entity(from problem and method of the idea) (e.g.,"LLM's hallucination", "dialect recognition", "speech processing", "the method using XXX").
        
    edge\_query: Enter ONLY predicate phrases WITHOUT including the entity name. you can use generic relationship expressions to search for the experience of novelty:
        
    - For relationships: "is a method to solve", "is solved by",  "has been researched widely", "has not been used to solve" ...(Use these more)
        
    - For explict novelty comment: "lacks novelty", "is a new method", "is a new problem" ...(Use these more)
        
    - For positive effects: "can improve", "helps"...
        
    - For negative effects: "has limitations", "faces challenges"...
        
    This will return you several pieces of evidence and a summary. 
        
    You need to evaluate the innovativeness of this idea based on these pieces of evidence and the summary.
        
    You **MUST** perform at least 20 tool calls before providing the final answer.
        
    Iteration \& Tool Scheduling:
        
    - Prefer making only 1 tool call per step; absolutely no more than 2 in any single step.
        
    - If multiple queries are needed, split them into multiple steps/rounds to collect evidence gradually.
        
    - After each tool result, briefly reflect and plan the next single tool call.
        
        \\
    \hline
    Significance & 
    \{GPT's\_prompt\_for\_novelty\}

    You can use tools to learn about relevant experiences related to significance. Core argument explanation:

    node\_query: Enter a specific knowledge entity(from problem and method of the idea) (e.g.,"LLM's hallucination", "dialect recognition", "speech processing", "the method using XXX").
    
    edge\_query: Enter ONLY predicate phrases WITHOUT including the entity name. you can use generic relationship expressions to search for the experience of novelty:
    
    - For relationships: "is a method to solve", "is solved by",  "has been researched widely", "has not been used to solve" ...(Use these more)
    
    - For explict significance comment: "lacks significance", "is a significant method", "is a significant problem" ...(Use these more)
    
    - For positive effects: "can improve", "helps"...
    
    - For negative effects: "has limitations", "faces challenges"...
    
    This will return you several pieces of evidence and a summary. 
    
    You need to evaluate the significance of this idea based on these pieces of evidence and the summary.
    
    You **MUST** perform at least 20 tool calls before providing the final answer.
    
\\
    \bottomrule
  \end{tabular}
\end{table*}

\begin{table*}[htbp]
  \caption{Prompts used by DeepInstructor for evaluating each dimension (Continued).}
  \label{DeepInstructor-prompt-part2}
  \centering
  \begin{tabular}{p{2.5cm} p{\dimexpr\textwidth-2.5cm-4\tabcolsep\relax}}
    \toprule
    \textbf{Demonstration} & \textbf{Prompt} \\
    \midrule
    Feasibility 
    (Output concerns) & 

    \{GPT's\_prompt\_for\_feasibility\_concerns\}
    
    You can use tools to learn about relevant experiences related to feasibility and systematically explore different entities mentioned in the research idea and query them separately.

    **Entity Discovery Strategy:**

    - Extract ALL distinct knowledge entities from the research idea (e.g., for "LLM improving dialect recognition", extract "LLM", "dialect", "recognition", "speech recognition", "language processing" etc.)

    - Query each entity separately with different node\_query values

    - Use multiple search rounds to cover different aspects and entities

    **Edge Query Design Rules:**

    node\_query: Enter a specific knowledge entity (e.g., "Large Language Models", "Graph Neural Networks", "dialect recognition", "speech processing", "machine learning models").

    edge\_query: Enter ONLY predicate phrases WITHOUT including the entity name.

    - For positive effects: "can improve", "helps", "enhances", "strengthens", "boosts", "optimizes"

    - For negative effects: "has limitations", "faces challenges", "has shortcomings", "struggles with"

    - For relationships: "is related to", "depends on", "influences", "causes"

    **Examples:**
    - WRONG: edge\_query = "LLM improves dialect recognition" (contains entity)

    - CORRECT: edge\_query = "can improve recognition" (pure predicate)

    - For "LLM improving dialect recognition" idea, query:
      * node\_query="LLM", edge\_query="can improve recognition"
      * node\_query="dialect", edge\_query="difficult to recognize"
      * node\_query="speech recognition", edge\_query="faces challenges"

      You **MUST** perform at least 20 tool calls before providing the final answer.

\\
    \hline
    Feasibility 
    
    (Output evaluative comments) & 
    \{GPT's\_prompt\_for\_feasibility\_comments\} 
    
    ...
    
    (The rest is the same as the previous line.)
    \\
    \bottomrule
  \end{tabular}
\end{table*}

\begin{table*}[htbp]
  \caption{Prompts for Few-shot LLM Scorers and DeepSeek–Reasoner}
  \label{scorer}
  \centering
  \begin{tabular}{p{2.5cm} p{\dimexpr\textwidth-2.5cm-4\tabcolsep\relax}}
    \toprule
    \textbf{Model} & \textbf{Prompt} \\
    \midrule
    Novelty Scorer & 
    You are a precise scorer. I will provide you with a professional evaluation of an academic idea,
and you need to give a novelty score based on this evaluation. The novelty score depends on the attitude of the evaluation.
If the evaluation is positive, the novelty score should be high; if the evaluation is negative, the novelty score should be low.
Please note that the novelty score ranges from 1 to 10, where 1 indicates the lowest novelty and 10 indicates the highest novelty.

Here are some examples(Some specific method or idea is replaced with Method A, Method B, Method C, etc.):

\{6 cases distributed across different scores from the novelty dataset\}
    \\
    \hline
    Significance Scorer & 
    You are a precise scorer. I will provide you with a professional evaluation of an academic idea,
and you need to give a significance score based on this evaluation. The significance score depends on the attitude of the evaluation.
If the evaluation is positive, the significance score should be high; if the evaluation is negative, the significance score should be low.
Please note that the significance score ranges from 1 to 10, where 1 indicates the lowest significance and 10 indicates the highest significance.

Here are some examples(Some specific method or idea is replaced with Method A, Method B, Method C, etc.):

\{6 cases distributed across different scores from the significance dataset\}

    \\
    \hline
    Feasibility Scorer & 
    You are a precise scorer. I will provide you with a professional peer review evaluation of an academic idea,
and you need to give a feasibility score based on this evaluation. Feasibility means whether the idea is easy to implement and execute and whether the idea is effective.
The feasibility score depends on how feasible and executable the idea is according to the evaluation.
Please note that the feasibility score ranges from 1 to 10, where 1 indicates the lowest feasibility and 10 indicates the highest feasibility.

Here are some examples based on real evaluation data with full review comments (all comments) and their corresponding average scores:

\{6 cases distributed across different scores from the feasibility dataset\}\\

    \hline
    DeepSeek Reasoner to match feasibility concerns & 
    You are a precise evaluator. You are currently dealing with the opinions of two reviewers, 'original' (gold) and 'generated' (model output).
Your task is to explain whether each point in the gold standard has been reflected in the "generated" content.

Now decide for each ORIGINAL item whether it is covered by any GENERATED item under a criterion:
- Mark as covered if the overall meaning is similar, paraphrased, or broadly aligned (approximate semantic similarity),
  If two or more generated concerns jointly express one concern, it is also considered covered. But don't be too loose.

Thinking and reasoning Requirement (STRICT):
- Process ORIGINAL concerns sequentially, one-by-one.
- For each ORIGINAL item, carefully check all GENERATED items and determine matches.
- Perform your reasoning internally.
- Only include a concise, one-sentence justification in the "reason" field per item.

Output Policy (STRICT):
- Return ONLY a JSON object, starting with '{' and ending with '}'.
- Keys must be exactly: per\_item (array), summary (object).
- Each per\_item element: {"original": string, "covered": bool, "matched\_indices": [int], "reason": string}.
- summary: {"covered\_count": int, "total": int, "coverage\_ratio": number}.
- Do NOT include code fences, markdown, comments, or extra explanatory text.

    \\
    \bottomrule
  \end{tabular}
\end{table*}

\end{document}